\documentclass[10pt,twocolumn,letterpaper]{article}

\usepackage[pagenumbers]{cvpr} % To force page numbers, e.g. for an arXiv version

\usepackage{graphicx}
\usepackage{amsmath}
\usepackage{amssymb}
\usepackage[accsupp]{axessibility} % Improves PDF readability for those with disabilities.

\usepackage{booktabs,multirow}
\usepackage{diagbox}
\usepackage{graphbox}
\usepackage{graphics}
\usepackage{xcolor}
\usepackage[super]{nth}
\newcommand{\stimes}{{\times}}
\usepackage{amsmath,amsfonts,bm}

\def\eqref#1{equation~\ref{#1}}
\def\1{\bm{1}}

\def\va{{\bm{a}}}
\def\vb{{\bm{b}}}
\def\vc{{\bm{c}}}
\def\vd{{\bm{d}}}

\def\vf{{\bm{f}}}

\def\vm{{\bm{m}}}

\def\vs{{\bm{s}}}

\def\mP{{\bm{P}}}

\DeclareMathAlphabet{\mathsfit}{\encodingdefault}{\sfdefault}{m}{sl}
\SetMathAlphabet{\mathsfit}{bold}{\encodingdefault}{\sfdefault}{bx}{n}
\newcommand{\tens}[1]{\bm{\mathsfit{#1}}}
\def\tA{{\tens{A}}}

\def\tD{{\tens{D}}}

\def\tH{{\tens{H}}}

\def\tL{{\tens{L}}}
\def\tM{{\tens{M}}}
\def\tN{{\tens{N}}}

\def\tP{{\tens{P}}}

\def\tR{{\tens{R}}}

\def\tU{{\tens{U}}}

\def\tH{h}
\def\tL{\ell}
\def\tM{m}
\def\tR{r}
\def\size1{1.364065}

\usepackage[pagebackref,breaklinks,colorlinks]{hyperref}

\usepackage[capitalize]{cleveref}
\crefname{section}{Sec.}{Secs.}
\Crefname{section}{Section}{Sections}
\Crefname{table}{Table}{Tables}
\crefname{table}{Tab.}{Tabs.}

\def\cvprPaperID{7722} % *** Enter the CVPR Paper ID here
\def\confName{CVPR}
\def\confYear{2022}

\begin{document}

\title{Image Animation with Perturbed Masks}

\author{Yoav Shalev \quad\quad\quad\quad\quad\quad\quad Lior Wolf \\
Blavatnik School of Computer Science, Tel Aviv University \\
{\tt\small yoavshalev@mail.tau.ac.il \quad wolf@cs.tau.ac.il}
\and
}
\maketitle

%%%%%%%%% ABSTRACT
\begin{abstract}
   We present a novel approach for image-animation of a source image by a driving video, both depicting the same type of object. We do not assume the existence of pose models and our method is able to animate arbitrary objects without the knowledge of the object's structure. Furthermore, both, the driving video and the source image are only seen during test-time. Our method is based on a shared mask generator, which separates the foreground object from its background, and captures the object's general pose and shape. To control the source of the identity of the output frame, we employ perturbations to interrupt the unwanted identity information on the driver's mask. A mask-refinement module then replaces the identity of the driver with the identity of the source. Conditioned on the source image, the transformed mask is then decoded by a multi-scale generator that renders a realistic image, in which the content of the source frame is animated by the pose in the driving video. Due to the lack of fully supervised data, we train on the task of reconstructing frames from the same video the source image is taken from. Our method is shown to greatly outperform the state-of-the-art methods on multiple benchmarks. Our code and samples are available at https://github.com/itsyoavshalev/Image-Animation-with-Perturbed-Masks.
\end{abstract}

%%%%%%%%% BODY TEXT
\section{Introduction}
\label{sec:introduction}

The ability to reanimate a still image based on a driving video has been extensively studied in recent years~\cite{wang2019fewshotvid2vid,ren2020deep,kim2018deep}. The developed methods achieve an increased degree of accuracy in both, maintaining the source identity, as extracted from the source frame, and in replicating the motion pattern of the driver’s frame. In addition, the recent methods also show good generalization to unseen identities and are relatively robust, and have fewer artifacts than the older methods. The relative ease with how these methods can be applied out-of-the-box has led to their adoption in various visual effects.

Interestingly, some of the most striking results have been obtained with model-free methods, i.e., that do not rely, for example, on post-extraction models~\cite{2016arXiv161107004I,wang2018pix2pixHD,2018arXiv180404732H,2018arXiv180710550W,Siarohin_2019_CVPR,siarohin2019first}. This indicates that such methods can convincingly disentangle shape and identity from motion~\cite{lorenz2019unsupervised,2020arXiv200109518D}.

There are, however, a few aspects in which such methods still need to improve. First, the generated videos are with noticeable artifacts. Second, some of the identity of the source image is lost and replaced by identity elements from the driving video. Third, the animation of the generated video does not always match the motion in the driver video.

Here, we propose a method that is preferable to the existing work in terms of motion accuracy, identity and background preservation, and quality of the generated video. Our method relies on a mask-based representation of the driving pose and explicit conditioning on the source foreground mask. Source and driver masks are extracted by the same network. The driver mask goes through an additional stage that replaces the identity information in the mask.

The reliance on masks has many advantages. First, it eliminates many of the identity cues from the driving video. Second, it explicitly models the region that needs to be replaced in the source image. Third, it is common to both source and driver, thus allowing, with proper augmentation, to train only on source videos. Fourth, it captures a detailed description of the object's pose and shape. 

Interestingly, unlike many of the previous methods, we do not rely on GANs~\cite{goodfellow2016deep} to generate proper outputs from combinations of different inputs. Instead, we employ an encoder-decoder, in which the identity is manipulated in order to direct the networks toward employing specific parts of the information from each input. To summarize, our contributions are: 
% (i) An image animation method that is based on applying a masking process to both the source image and the driving video;
% (ii) The method generalizes to unseen identities of the same type and is able to animate arbitrary objects better than previous work;
% (iii) The mask generator separates the foreground object from its background and captures, in a generic way, the fine details of the object's pose and shape;
% (iv)  Conditioning the mask of the driving frame on the source frame and its mask to introduce the identity of the source frame; 
% (v) Employing perturbations in order to remove the foreground identity of the driver's frame from the mask; and
% (vi) A comprehensive evaluation of several different applications of our method, which show a sizable improvement over the current state of the art in image animation.
{
(i) An image animation method that generalizes to unseen identities of the same type, and is able to animate arbitrary objects better than previous work;
(ii) Innovative use of perturbations over masks, in order to interrupt the driver's identity, which is then replaced with the source's identity by the mask refinement module; 
(iii) A comprehensive evaluation of several different applications, which show a sizable improvement over the current image animation state-of-the-art.
}
%\end{tight_itemize}

%-------------------------------------------------------------------------
\section{Related Work}
\label{sec:relatedwork}

Much of the work on image animation relies on prior information on the animated object, in the form of explicit modeling of the object's structure, e.g., some methods animate a source image using facial landmarks~\cite{zakharov2019few, 2020arXiv200810174Z}, while~\cite{ren2020deep} developed a human-pose-guided image generator. However, in many applications, an explicit model is not available. Our method is model-free and able to animate arbitrary objects.

There are many model-free contributions in the field of image-to-image translation, where an image of one domain is mapped to an analog image of another domain. \cite{2016arXiv161107004I} learns a map between two domains using a conditional GAN. \cite{wang2018pix2pixHD} developed a multi-scale GAN that generates high-resolution images from semantic label maps. \cite{2018arXiv180404732H} encodes images of both domains into a shared content space and a domain-specific style space. The content code of one domain is combined with the style code of the other domain, and then the image is generated using a domain-specific decoder. For this class of methods, the model is not able to generalize to other unseen domains of the same category without retraining. In contrast, for a given type of model (e.g. faces), our method is trained once, and able to generalize to unseen domains of the same type (e.g. the source and driving faces can be of any identity).

More related to our method is a method that assumes a reference frame for each video, and learns a dense motion field that maps pixels from a source frame to its reference frame, and another mapping from the reference frame to the driver's frame \cite{2018arXiv180710550W}. ~\cite{Siarohin_2019_CVPR} extracts landmarks for driving and source images of arbitrary objects, and generates motion heatmaps from the key-points displacements. The heatmaps and the source image are then processed to generate the final prediction. 

A follow-up work~\cite{siarohin2019first} extracts a first-order motion representation, consisting of sparse key-points and local affine transformations, with respect to a reference frame. The motion representation is then processed to generate a dense motion field, from the driver's frame to the source's, and occlusion maps to mask out regions that should be inpainted by the generator. {This method, like ours, does not employ GANs. The main differences are that our method does not assume a reference frame, instead of key-points, we generate objects masks, which are more informative regarding pose and shape, and our innovative use of perturbations and the mask refinement module.}

Other methods, including \cite{lorenz2019unsupervised,2020arXiv200109518D}, learn a part-based disentangled representation of shape and appearance, and try to ensure that local changes in appearance and shape remain local, and do not affect the overall representation. On the other hand, our method does not assume a predefined number of parts, and by using perturbations and the mask refinement module, it is able to better remove the driver’s identity and inject that of the source.

When a source video is available, video-to-video translation methods~\cite{2018arXiv180807371C,kim2018deep} may be used for motion transfer, by utilizing the rich appearance and pose information of the source video. Such methods learn a mapping between two domains and are able to generate realistic results, where the source video is animated by the driver video. These methods require a large number of source frames at train time, and require a long training process for every target subject. In contrast, our model is able to animate a single source image, which is unseen during training, and employs a driving video with another novel person. %and it's generalizes to unseen source and driving objects of the same category. % training process for each domain is time consuming. In addition, it requires a separate generator for each subject.

\begin{figure*}[t]
\centering
\begin{tabular}{cc}
\includegraphics[width=.3\linewidth, height=45mm]{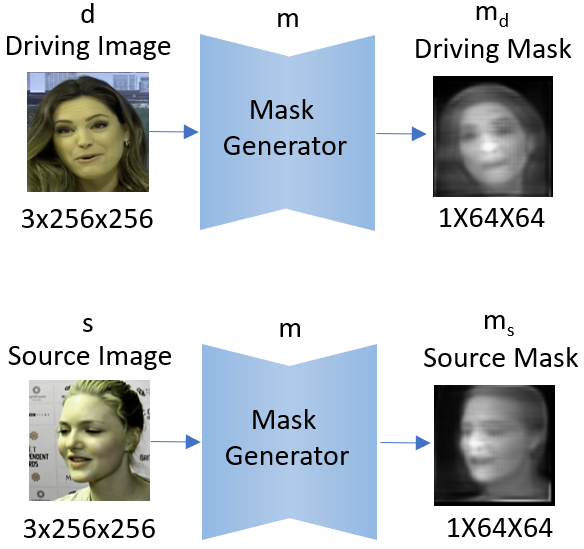} & \includegraphics[width=.5\linewidth, height=45mm]{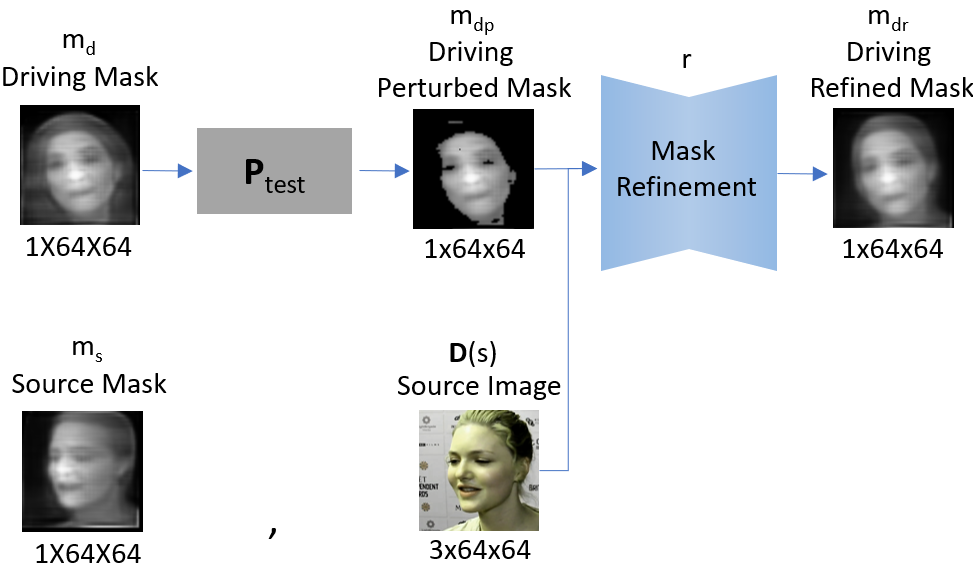} \\
(a) mask generator $\tM$ & (b) mask refinement network $\tR$ \\ \\
\includegraphics[width=0.5\linewidth, height=45mm]{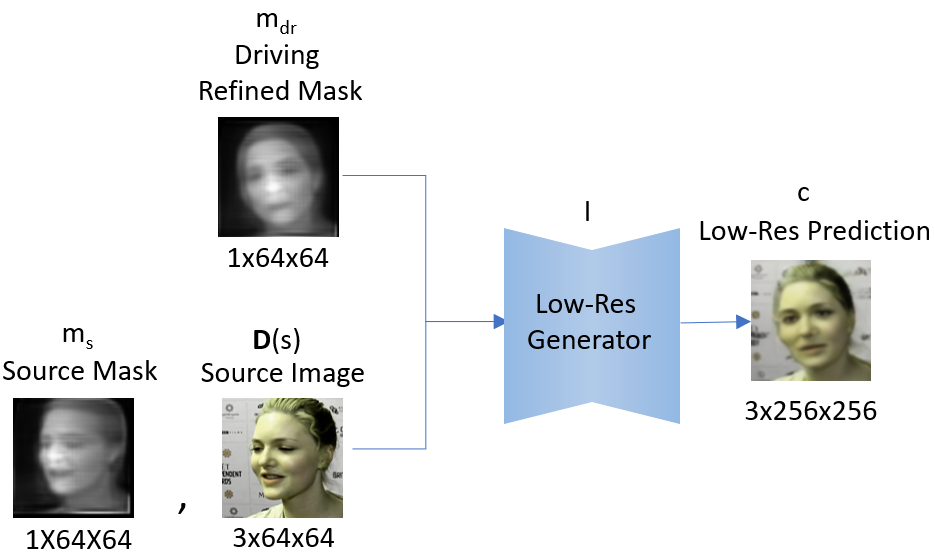} &   \includegraphics[width=0.4\linewidth, height=45mm]{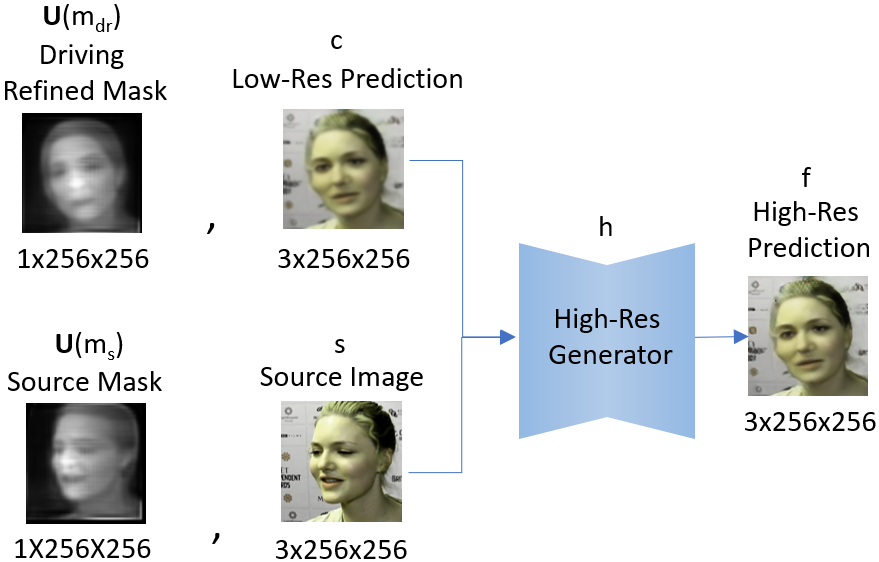} \\
(c) low-res generator $\tL$ & (d) high-res generator $\tH$
% \multicolumn{2}{c}{\includegraphics[width=65mm]{it} }\\
% \multicolumn{2}{c}{(e) fifth}
\end{tabular}
\caption{Overview of our method at test time. (a) Source and driving masks $\vm_s$ and $\vm_d$ are generated using the mask generator $\tM$. (b) The identity-perturbation operator $\tP_\text{test}$ is then applied to the driver's mask, and along with a scaled-down version of the source's image $\tD(\vs)$ and the source's mask $\vm_s$, they are fed into the mask refinement network $\tR$, to generate the driver's refined mask $\vm_{dr}$. (c) The refined mask $\vm_{dr}$, the source's mask $\vm_s$, and the scaled-down source's image $\tD(\vs)$ are fed into the generator $\tL$, which generate the initial prediction $\vc$. (d) The scaled-up refined mask $\tU(\vm_{dr})$, the source image $\vs$, the initial prediction $\vc$, and the scaled-up source's mask $\tU(\vm_s)$ are fed into the generator $\tH$, in order to generate the final prediction $\vf$.}
\label{fig:test_figure}
\end{figure*}

\section{Method}
\label{sec:method}

The method consists of four {encoder-decoder networks}: the mask generator $\tM$, the mask refinement network $\tR$, and the low and high-resolution frame generators $\tL$ and $\tH$. The networks transform a source frame $\vs$ and a driving frame $\vd$ into the generated high-resolution frame $\vf$, {{where $\vf$ contains the foreground and background of the source frame $\vs$, such that the pose of the foreground object in $\vs$ is modified to match that of the driver frame $\vd$. This is done for each driving frame separately, and at test-time executed through the following process, as depicted in Fig.~\ref{fig:test_figure}}}:
% \begin{align}
% \vm_s &= \tM(\vs) \label{eq:m1}\\
% \vm_d &= \tR(\tD(\vs), \vm_s, \tP_\text{test}(\tM(\vd))) \label{eq:m2}\\
% \vc &= \tL(\tD(\vs),\vm_s,\vm_d)\label{eq:c}\\
% \vf &= \tH(\vs,\tU(\vm_s),\tU(\vm_d),\vc)\label{eq:f}\,,
% \end{align}
\begin{align}
\vm_s &= \tM(\vs) \label{eq:m1}\\
\vm_d &= \tM(\vd) \label{eq:m2}\\
\vm_{dp} &= \tP_\text{test}(\vm_d)\label{eq:m3}\\
\vm_{dr} &= \tR(\tD(\vs), \vm_s, \vm_{dp}) \label{eq:m4}\\
\vc &= \tL(\tD(\vs),\vm_s,\vm_{dr})\label{eq:c}\\
\vf &= \tH(\vs,\tU(\vm_s),\tU(\vm_{dr}),\vc)\label{eq:f}\,,
\end{align}

\noindent where upper-cased bolded notations represent untrained operations, including $\tD$ ($\tU$), which is a downscale (upscale) operator, implemented using a bi-linear interpolation, that transforms an image of resolution $256 \times 256$ to an image of resolution $64 \times 64$ (or vice versa).  

First, $\vm_s$ and $\vm_d$ are generated using the mask generator $\tM$.
Next, the identity-perturbation operator $\tP_\text{test}$ is applied on the driver's mask $\vm_d$, by setting to zero pixels that are smaller than a threshold $\rho$. Considering typical face masks, e.g., the pixels in the areas of the eyes, mouth, and hair are with low intensities. Removing these pixels by applying $\tP_\text{test}$, results in a much more generic face, interrupting the driver's identity. For each driver's mask, we set the threshold $\rho$ to be the median pixel value.

Next, the refinement network $\tR$ acts to generate the missing data of the perturbed mask $\vm_{dp}$ and to replace the driver's identity with that of the source. It uses the source's frame and mask as a reference.

Finally, the generated frame is being synthesized in a hierarchical process in which the coarse (low resolution) frame $\vc$ is first generated using $\tL$ and is then refined by the network $\tH$. Both generators ($\tL,\tH$) utilize the mask $\vm_s$ to attend the foreground and background objects in the source {frame} $\vs$, and to infer the occluded regions that need to be generated.

The refined driver's mask $\vm_{dr}$ is the only conditioning on the frame generation process that stems from the driver's frame $\vd$. It, therefore, needs to encode the pose of the foreground object in the driving frame. However, this has to be done in a way that is invariant to the driver's identity. For example, when reanimating person A based on a driver video of person B, the pose of B should be given, while discarding the body shape information of B. Otherwise, the generated frame could have the appearance of the source's foreground and a body shape that mixes that of the person in the source frame and that of the person in the driving frame.  {The perturbation operator $\tP_\text{test}$ is, therefore, designed to interrupt the elements that are associated with the driver's identity, which encourages the refinement network $\tR$ to project identity elements from the reference mask ($\vm_s$) and frame ($\vs$). As a result, the proposed identity replacement stage does not modify the general pose of the driver's mask, but only replaces the driver's identity.}

\subsection{Training}

\begin{figure*}
\centering
\begin{tabular}{cc}
% \includegraphics[width=.3\linewidth, height=40mm]{Overview/train_mask_gen.PNG} & \includegraphics[width=.5\linewidth, height=40mm]{Overview/train_mask_ref.PNG} \\
% (a) mask generator $\tM$ & (b) mask refinement network $\tR$ \\ \\
\includegraphics[width=0.5\linewidth, height=50mm]{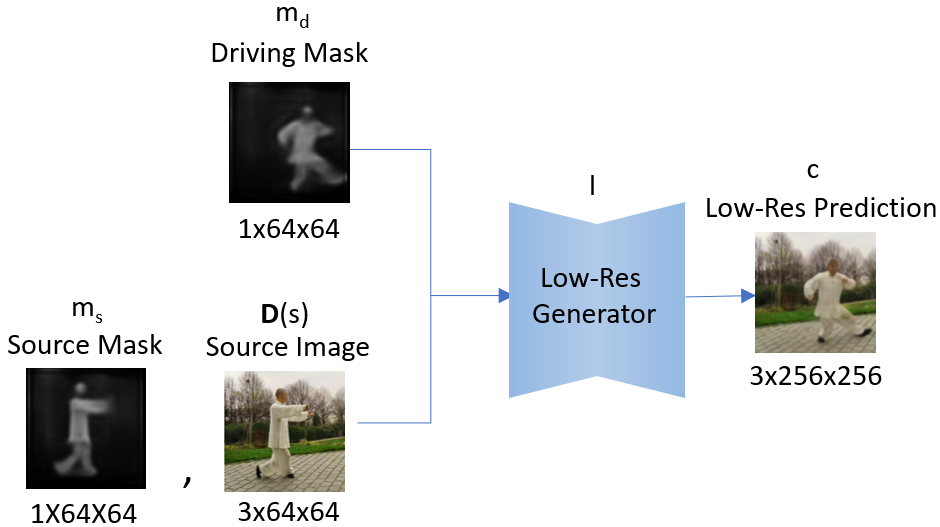} &   \includegraphics[width=0.5\linewidth, height=50mm]{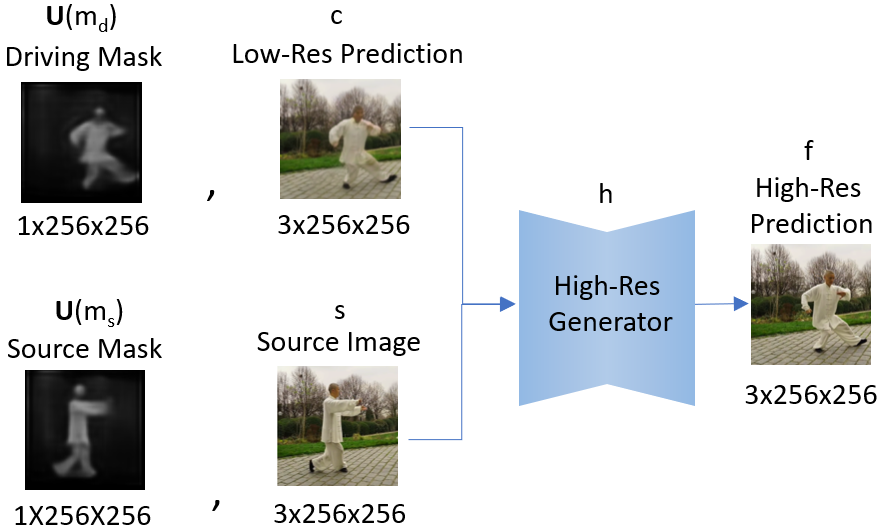} \\
(a) low-res generator $\tL$ & (b) high-res generator $\tH$
% \multicolumn{2}{c}{\includegraphics[width=65mm]{it} }\\
% \multicolumn{2}{c}{(e) fifth}
\end{tabular}
\caption{The low-res and high-res generators at train time. Instead of getting the driver's refined mask $\vm_{dr}$ as in test time, the two generators $\tL$ and $\tH$ are using the driver’s mask $\tM(\tA(\vd))$ and its up-scaled version $\tU(\tM(\tA(\vd)))$, respectively.}
\label{fig:train_figure}
\end{figure*}

Training is conducted using driving and source frames from the same video. The reason is that for the type of supervised loss terms we use, a ground-truth target frame is required. The main challenge is to keep the model robust enough for accepting at test time a driving frame $\vd$ from another video. 

The training pipeline is slightly modified from test time, in that an augmentation $\tA$ is applied to the driving frame $\vd$, and that a more elaborate perturbation $\tP_\text{train}$ takes a place. In addition, since the source and driving frames are of the same identity, as shown in Fig.~\ref{fig:train_figure} both generators $\tL$ and $\tH$ are using the driver’s mask $\vm_d$, instead of using the refined mask $\vm_{dr}$, which is used only for training the refinement network $\tR$:
\begin{align}
\vm_d &= \tM(\tA(\vd)) \label{eq:train_m1}\\
\vm_{dp} &= \tP_\text{train}(\vm_d) \label{eq:train_m2}\\
\vm_{dr} &= \tR(\tD(\vs), \vm_s, \vm_{dp}) \label{eq:train_m3}\\
\vc &= \tL(\tD(\vs),\vm_s,\vm_d)\label{eq:train_c}\\
\vf &= \tH(\vs,\tU(\vm_s),\tU(\vm_d),\vc)\label{eq:train_f}\,,
\end{align}

\noindent where augmentation $\tA$ is a color transformation that scales the input's brightness, contrast, and saturation by a random value drawn from $[0.9, 1.1]$, and shifts its hue by a random value drawn from $[-0.1, 0.1]$. {The goal of this augmentation is to encourage the generated masks to be invariant to the input's appearance, despite the challenge mentioned above of training on frames from the same video.}

$\mP_\text{train}$ performs the following two steps sequentially: (i) breaks the image vertically (horizontally) into six parts, and scales each part horizontally (vertically) by a random value drawn from $[0.75, 1.25]$. Next, it scales the entire output vertically (horizontally), by a random value drawn from $[0.75, 1.25]$. (ii) {similarly to}  $\mP_\text{test}$, sets to zero pixels that are smaller than a threshold value $\rho$, which we set to be the median pixel value of each mask. %and, finally (iii) adds an element-wise noise sampled from the Poisson distributions with $\lambda=20$. 
The goal of the resizing operation is to interrupt the driver's identity by modifying the proportions of its mask, e.g., in faces, it modifies the distance between the eyes, which results in an identity perturbation, while keeping the general pose. The thresholding 
%and noising operations 
operation eliminates low-intensity pixels (e.g. the boundary of the body, and the hair, eyes, and mouth areas), which are a major ingredient of the driver's identity. 
Without the listed operations of $\mP_\text{train}$, we experienced a phenomenon where the mask refinement module $\tR$ ignores the reference mask ($\vm_s$) and frame ($\vs$), i.e. applying $\mP_\text{train}$ encourage the mask refinement network $\tR$ to project the elements that are associated with the source’s identity, which is crucial for the generation part.

All hyper-parameters values, including these constants, were selected using cross-validation %on the training set of the Tai-Chi~\cite{siarohin2019first} dataset
and fixed throughout all experiments on all benchmarks.

\smallskip
\noindent{\bf Loss Terms\quad} 
The model is trained end-to-end using two loss terms: a mask refinement loss and a perceptual reconstruction loss. At train time, where the source and driving frames are of the same identity, the role of the mask refinement network $\tR$ is to recover the missing data that was removed by the operator $\tP_{\text{train}}$. Therefore, we minimize the ${L}_1$ loss of the driver's mask $\vm_d$ and its refined mask $\vm_{dr}$:
\begin{equation}
\mathcal{L}_{\text{mask}}(\vd) = {L}_1({\vm_{dr}}, \vm_d)\,.
\end{equation}
%where $\beta$ is an hyper-parameter set to $100$.

For the image reconstruction loss of the generators $\tL$ and $\tH$, following \cite{siarohin2019first} and based on the implementation of \cite{2018arXiv180806601W}, we minimize a perceptual loss using the pre-trained weights of a VGG-19 model. For two images $\va$ and $\vb$, the reconstruction loss terms using the $j^{th}$ layer of the pre-trained VGG model are written as:
\begin{align}
\mathcal{L}_{\text{VGG}}(\va, \vb)_j = \text{AVG}(|\tN_{j}(\va)-\tN_{j}(\vb)|)
\end{align}
% \mathcal{L}_{coarse}(c, d)_j = \lambda_j \cdot \frac{\sum_{i=1}^{I_j} |N_{ji}(c)-N_{ji}(d)|}{I_j}\\
% \mathcal{L}_{fine}(f, d)_j = \lambda_j \cdot \frac{\sum_{i=1}^{I_j} |N_{ji}(f)-N_{ji}(d)|}{I_j}
% \end{align}
where $\text{AVG}$ is the average operator and $\tN_{j}(\cdot)$ are the features extracted using the $j^{th}$-layer of the pre-trained VGG model. For the coarse and fine predictions $\vc$ and $\vf$, and a driving frame $\vd$, we compute the following reconstruction loss for multiple resolutions:
\begin{align}
    \mathcal{L}_{\text{reconstruct}} =  \sum_s \sum_j  \mathcal{L}_{\text{VGG}}(\vc_{\vs}, \vd_{\vs})_j + \mathcal{L}_{\text{VGG}}(\vf_{\vs}, \vd_{\vs})_j\notag
\end{align}
where input image $\va_{\vs}$, has a resolution $\vs \in [256^2,128^2,64^2]$. We use the first, third, and fifth ReLU layers of the VGG-19 model. Note that while VGG was designed for a resolution of $224^2$, the first layers are convolutional, and can be used for an arbitrary input scale.

The combined loss is given by $\mathcal{L}=\lambda_1\mathcal{L}_\text{mask} + \lambda_2\mathcal{L}_\text{reconstruct}$, for weight parameters $\lambda_1=100$ and $\lambda_2=10$. To avoid unwanted adaptation of the network $\tM$, the backpropagation of $\mathcal{L}_\text{mask}$ only updates the weights of the mask refinement network $\tR$. When backpropagating the second part of the reconstruction loss $ \sum_s \sum_j  \mathcal{L}_{\text{VGG}}(\vf_{\vs}, \vd_{\vs})_j$, only the generator $\tH$ is updated. The Adam optimizer is employed with a learning rate of $2 \times 10 ^{-4}$ and $\beta$ values of $0.5$ and $0.9$. The batch size is $16$. Following~\cite{siarohin2019first}, we decay the learning rate at epochs $60$ and $90$, running for $100$ epochs on NVIDIA Titan RTX. The mask refinement network $\tR$ starts training after we complete the first training epoch, when the outputs of the mask generator $\tM$ start to be meaningful.

The architecture of the networks is given in the supplementary materials, which also contain the source code.

\section{Experiments}
\label{sec:experiments}

The training and evaluation were done using three different datasets, containing short videos of diverse objects. \textbf{Tai-Chi-HD} is a dataset containing videos of people doing tai-chi exercises. Following \cite{siarohin2019first}, 3,141 tai-chi videos were downloaded from YouTube. The videos were cropped and resized to a resolution of $256^2$, while preserving the aspect ratio. There are 3,016 training videos and 125 test videos. \textbf{VoxCeleb} is an audio-visual dataset consist of short videos of talking faces, introduced by \cite{Nagrani17}. {VoxCeleb1} is the collection used, and as pre-processing{,} bounding boxes of the faces were extracted and resized to $256^2$, while preserving the aspect ratio. It contains an overall number of 18,556 training videos and 485 test videos. The {\textbf{BAIR} dataset contains videos of Sawyer robotic arms interacting with objects \cite{2017arXiv171005268E}.} It contains 42,880 training videos and 128 test videos, where each video consists of 30 frames with a resolution of $256^2$. We were unable to obtain the UvA-NEMO dataset \cite{dibekliouglu2012you}, which was utilized in some earlier contributions.

%In order to evaluate our method and compare to the state of the art, 
{ We borrow and significantly expand the evaluation process of \cite{siarohin2019first}. Our method is evaluated quantitatively and qualitatively for the tasks of both video reconstruction and image animation, where the source and driving videos are of different identities.} Additionally, despite being model-free, we compare to model-based methods in the few-shot-learning scenario. In this case, our method, unlike the baseline methods, does not employ any few shot samples. %Finally we compare our method with previous works in the field of model-free image animation.

% previous work, %the evaluation process of \cite{Siarohin_2019_CVPR,zakharov2019few}, 
Multiple metrics are used for evaluation: \textbf{L1} is the L1 distance between the generated and ground-truth videos. \textbf{Average Key-points Distance (AKD)} measures the average distance between the key-points of the generated and ground-truth videos. For {Tai-Chi-HD}, we use the human-pose estimator of \cite{2016arXiv161108050C}, and for VoxCeleb, we use the facial landmark detector of \cite{2017arXiv170307332B}. \textbf{Missing Key-points Rate (MKR)} measures the percentage of key-points that were successfully detected in the ground-truth video, but were missing in the generated video. The human-pose estimator of \cite{2016arXiv161108050C} outputs for every keypoint an indicator of whether it was successfully detected. Using this indicator, we measure MKR for the {Tai-Chi-HD} dataset. \textbf{Average Euclidean Distance (AED)} measures the average Euclidean distance in some embedding space between the representations of the ground-truth and generated videos. Following \cite{siarohin2019first}, we employ the feature embedding of \cite{Siarohin_2019_CVPR}. \textbf{Structural Similarity (SSIM)} \cite{wang2004image}: %a measure for image quality degradation; %following \cite{zakharov2019few}, 
For VoxCeleb, we compare the structural similarity of the ground-truth driving frames and generated images. \textbf{Cosine Similarity (CSIM)}: For VoxCeleb, we measure the identity similarity of the generated and ground-truth source faces, by comparing the cosine similarity of embedding vectors generated by a face recognition network \cite{deng2019arcface}. \textbf{Classification (CLS)}: For {Tai-Chi-HD}, we classify the generated frames using the Detectron2 framework \cite{wu2019detectron2}, and measure the number of frames classified as a person. Specifically, we use the X101-FPN COCO instance-segmentation model. \textbf{Intersection Over Union (IOU)}: For {Tai-Chi-HD}, we calculate the IOU of the segmentations of the generated and driving videos. The segmentations are generated using the same model we use for classification. \textbf{Facial Expression Similarity (FES)}: For VoxCeleb, we measure the facial expression similarity of generated and driving frames using the FER classifier (\url{https://github.com/justinshenk/fer}), which supports seven different emotions.
% \textbf{Fréchet inception distance (FID)}: FID is a similarity measurement between two sets of images, and is often used to evaluate the quality of generated images. The FID is calculated by computing the Fréchet distance between two Gaussians fitted to feature representations of the Inception network. We use the implementation of \cite{Seitzer2020FID}.

\subsection{Video Reconstruction}

The video reconstruction benchmarks follow the training procedure in that the source and target frames are from the same video. For evaluation, the first frame of a test video is used as the source frame, and the remaining frames of the same video as the driving frames. The goal is to reconstruct all the frames of the test video, except the first.

L1, AKD, MKR, and AED are compared with the state-of-the-art model-free methods, including X2Face of \cite{2018arXiv180710550W}, MonkeyNet of \cite{Siarohin_2019_CVPR}, and the method suggested by \cite{siarohin2019first}, which we refer to as FOMM. The results are reported in Tab.~\ref{quantitative_results_table}. Evidently, our method outperforms the baselines for each of the datasets and all metrics by a significant margin, except for the AKD measure on the VoxCeleb dataset, where accuracy was decreased by $2.7\%$. The most significant improvement is for the Tai-Chi-HD dataset, which is the most challenging dataset, because it consists of diverse movements of a highly non-rigid body.

Next, we follow \cite{zakharov2019few} and compare SSIM and CSIM with X2Face, Pix2PixHD \cite{wang2018high}, and the FSAL method \cite{zakharov2019few}. The baselines are evaluated in the few-shot-learning setting, where models are fine-tuned on a set of size \#FT, consisting of frames of a person that was not seen during the initial meta-learning step.  After the fine-tuning step, the evaluation is done on a hold-out set, consisting of unseen frames of the same person. The evaluation is done for VoxCeleb and the results are reported in Tab.~\ref{few_shots_table}. As can be seen, our method generalizes better and outperforms the baselines in SSIM and even more so in CSIM. This is especially indicative of the method's capabilities, since (i) we skip the fine-tuning step for our model (in our case $\text{\#FT}=0$), and (ii)  X2Face and FSAL were designed specifically for faces, while our method is model-free and generic.

\begin{table}[t]
\centering
\resizebox{0.8\linewidth}{!}{
\begin{tabular}{@{}l@{~}c@{~}c@{~}c@{~}c@{~}c@{~}c@{~}c@{~}c@{}}
\toprule
&\multicolumn{4}{c}{\textit{Tai-Chi-HD}} 
&\multicolumn{3}{c}{\textit{VoxCeleb}} 
&\multicolumn{1}{c}{\textit{BAIR}}\\
\cmidrule(lr){2-5}
\cmidrule(lr){6-8}
\cmidrule(lr){9-9}
{Method} 
&\multicolumn{1}{c}{L1}
&\multicolumn{1}{c}{AKD}
&\multicolumn{1}{c}{MKR}
&\multicolumn{1}{c}{AED} 
&\multicolumn{1}{c}{L1}
&\multicolumn{1}{c}{AKD}
&\multicolumn{1}{c}{AED} 
&\multicolumn{1}{c}{L1}\\
\midrule
X2Face & 0.080 & 17.654 & 0.109 & 0.272 & 0.078 & 7.687 & 0.405 & 0.065 \\
MN  & 0.077 & 10.798 & 0.059 & 0.228 & 0.049 & 1.878 & 0.199 & 0.034
\\
FOMM & 0.063 & 6.862 & 0.036 & 0.179 & 0.043 & \bf 1.294 & 0.140 & 0.027
\\
Ours &  \bf 0.047 & \bf 4.239 & \bf 0.015 & \bf 0.147 & \bf 0.034 & 1.329 & \bf 0.130 & \bf 0.021\\
\bottomrule
\end{tabular}}
%\smallskip %endofcenter
\caption{Video reconstruction results. MN=Monkey-Net.}
\label{quantitative_results_table}
\end{table}

\begin{table}[t]
\centering
\resizebox{0.8\linewidth}{!}{
\begin{tabular}{lccc}%{@{}l@{~~}c@{~~}c@{~~}c@{}}
\toprule
{Method}& \#FT
& SSIM $\uparrow$
& CSIM $\uparrow$
\\ \midrule
X2Face & 1/8/32 & 0.68/0.73/0.75 & 0.16/0.17/0.18
\\
P2PHD & 1/8/32 & 0.56/0.64/0.70 & 0.09/0.12/0.16
\\
FSAL& 1/8/32 & 0.67/0.71/0.74 & 0.15/0.17/0.19
\\ \midrule
% X2Face & 1 & 0.68 & 0.16
% \\
% Pix2PixHD & 1 & 0.56 & 0.09
% \\
% FSAL& 1 & 0.67 & 0.15
% \\ \midrule
% X2Face & 8 & 0.73 & 0.17
% \\
% Pix2PixHD & 8 & 0.64 & 0.12
% \\
% FSAL & 8 & 0.71 & 0.17
% \\ \midrule
% X2Face & 32 & 0.75 & 0.18
% \\
% Pix2PixHD & 32 & 0.70 & 0.16
% \\
% FSAL & 32 & 0.74 & 0.19
% \\ \midrule
Ours & {\bf 0} & \textbf{0.80} & \textbf{0.70}
\\
\bottomrule
\end{tabular}}
%\smallskip %endofcenter 
\caption{Few-shot learning results for VoxCeleb. Unlike baselines, we do not perform identity fine-tuning. \#FT=number of {frames} used for finetuning. P2PHD=Pix2PixHD.}
\label{few_shots_table}
%\end{table}
\end{table}

% \begin{table*}[t]
% \centering
% %\resizebox{1.0\linewidth}{!}{
% {
% \begin{tabular}{@{}l@{~}c@{~}c@{~}c@{~}c@{~}c@{~}}
% \toprule
% &\multicolumn{4}{c}{\textit{Tai-Chi-HD}} 
% &\multicolumn{1}{c}{\textit{VoxCeleb}} \\
% \cmidrule(lr){2-5}
% \cmidrule(lr){6-6}
% {Method} 
% &\multicolumn{1}{c}{L1}
% &\multicolumn{1}{c}{AKD}
% &\multicolumn{1}{c}{MKR}
% &\multicolumn{1}{c}{AED} 
% &\multicolumn{1}{c}{L1} \\
% \midrule
% FOMM & 0.068 & 8.561 & 0.043 & 0.196 & 0.050 \\
% Ours &  \bf 0.047 & \bf 4.239 & \bf 0.015 & \bf 0.147 & \bf 0.034 \\
% \bottomrule
% \end{tabular}
% }
% \smallskip
% \caption{{Video reconstruction using a wider bottleneck for baselines.}}
% \label{anim_quan_table}
% \end{table*}

\subsection{Image Animation}

The task of image animation is to animate a source image using a driving video. The object and its background in the source and driving inputs may have different identities and appearances. In the experiments, the first frame of a source video is used for encoding the appearance, and all frames of the driving video are used for driving the object's motion. A video is generated where the content of the source frame is animated by the driving video.

To evaluate the alignment between the generated and driving videos, we measure AKD, MKR, and IOU for the Tai-Chi-HD dataset, and FES and CSIM for the VoxCeleb dataset. AKD, MKR, and IOU are irrelevant for the VoxCeleb dataset, because a perfect match may indicate an identity loss. The reason is that the facial key-points and segmentations of different people have different ratios, and therefore cannot be compared. This is not the case for the Tai-Chi-HD dataset, where the camera is far from the person, and the body proportions are almost identical across different identities. Measuring CLS for the Tai-Chi-HD dataset provides differentiation, while for VoxCeleb, our method and FOMM are both almost 100\% accurate, and the improvement we present is negligible. Measurements are not available for the Bair dataset, due to the lack of a pre-trained classifier and a keypoint detector for the Sawyer robotic arm. For the following experiments, 100 pairs with different identities were randomly selected from the test set of each dataset. The quantitative animation results are presented in Tab.~\ref{anim_quan_table} and in Tab.~\ref{CSIM_anim_table}. As can be seen, our method is better for all metrics by a significant margin. See CSIM analysis for the ablation models in section ~\ref{ablation_sec}.

To evaluate the robustness for different levels of changes in pose, between source and driving frames, we extend the AKD, MKR, and FES experiments. Based on the AKD score between source and driving frames, we split the test set into three sub-sets, where the first sub-set contains the frames with the lowest score, and so on. We compare to FOMM, the most competitive method, and report the results in Tab.~\ref{groups_tab}. As can be seen, our method better preserves the driver's pose and expression, even for large changes.

% AKD, MKR, CLS and IOU are compared with the baselines for Tai-Chi-HD dataset. Note that comparing the AKD for the VoxCeleb dataset is irrelevant, because a perfect match may indicate an identity loss. The reason is that the facial key points of different people have different ratios, and therefore cannot be compared. This is not the case for the Tai-Chi-HD dataset, where the camera is far from the person, and the body proportions are almost identical.

% To calculate the FID, it is required that the number of samples in the dataset is greater than the dimension of the coding layer. Since some of the videos are very short, we use the first max-pooling features layer of the inception network, which has a dimension of 64. For the Tai-Chi-HD and VoxCeleb datasets, we treat each video as a separate dataset, and average the FIDs of generated and source videos. Most of the videos in the BAIR dataset are shorter than 64 frames, and since all videos contain the same Sawyer robotic arm, the FID was calculated once between all generated and source frames. 

\begin{figure}[t!]
\centering
\resizebox{0.85\linewidth}{!}{
%\framebox[4.0in]{$\;$}
%\begin{tabular}{@{}l@{}c@{}c@{}c@{}c@{}c@{}c@{}}
\begin{tabular}{@{}l@{}c@{}c@{}c@{}c@{}c@{}}
\toprule
 \parbox[t][][t]{1.2cm}{Vox-\\Celeb}
 &{\diagbox[height = 1.3cm,width=1.3cm]{\footnotesize{Source}}{\footnotesize{Driver}}} %\vline
&{\includegraphics[align=c,  height=1.3cm,width=1.3cm]{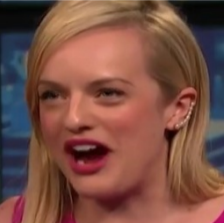}}
&{\includegraphics[align=c,  height=1.3cm,width=1.3cm]{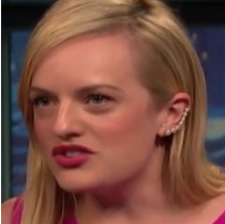}}
&{\includegraphics[align=c,  height=1.3cm,width=1.3cm]{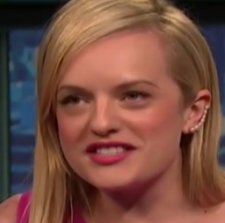}}
&{\includegraphics[align=c,  height=1.3cm,width=1.3cm]{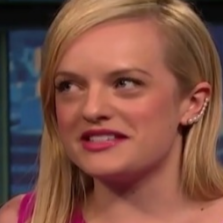}}
%&{\includegraphics[align=c,  height=1.3cm,width=1.3cm]{image_animation_results/vox/d_0000004.png}}
\\ %\cmidrule{1-7}
%\multirow{11}{*}{\textit{\rotatebox{90}{VoxCeleb}}} 
 X2Face & \includegraphics[align=c, height=1.3cm,width=1.3cm]{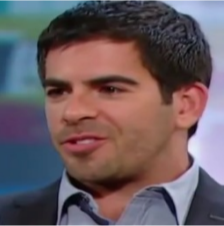} & \includegraphics[align=c, height=1.3cm,width=1.3cm]{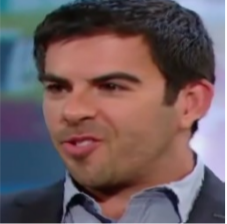} & \includegraphics[align=c, height=1.3cm,width=1.3cm]{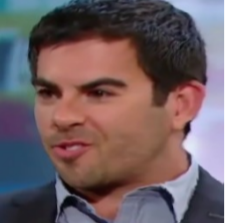} & \includegraphics[align=c, height=1.3cm,width=1.3cm]{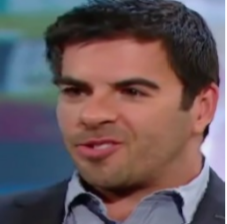} & \includegraphics[align=c, height=1.3cm,width=1.3cm]{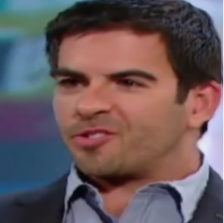} %&\includegraphics[align=c, height=1.3cm,width=1.3cm]{image_animation_results/vox/x2face/g_0000004.png} 
 \\
  MkeyN & \includegraphics[align=c, height=1.3cm,width=1.3cm]{image_animation_results/vox/source.png} & \includegraphics[align=c, height=1.3cm,width=1.3cm]{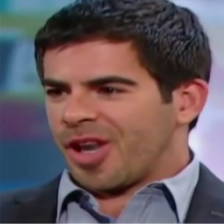} & \includegraphics[align=c, height=1.3cm,width=1.3cm]{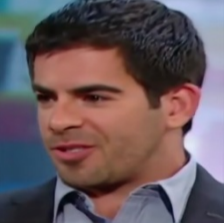} & \includegraphics[align=c, height=1.3cm,width=1.3cm]{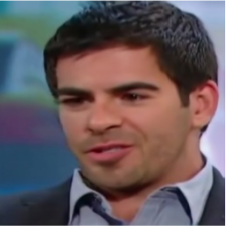} & \includegraphics[align=c, height=1.3cm,width=1.3cm]{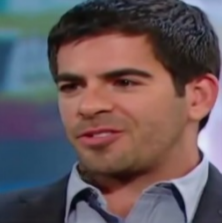} %&\includegraphics[align=c, height=1.3cm,width=1.3cm]{image_animation_results/vox/mn/g_0000004.png} 
  \\
  FOMM & \includegraphics[align=c, height=1.3cm,width=1.3cm]{image_animation_results/vox/source.png} & \includegraphics[align=c, height=1.3cm,width=1.3cm]{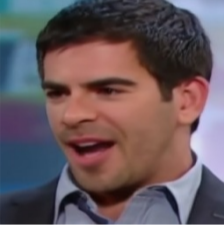} & \includegraphics[align=c, height=1.3cm,width=1.3cm]{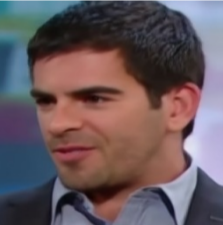} & \includegraphics[align=c, height=1.3cm,width=1.3cm]{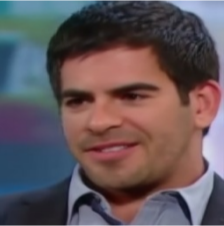} & \includegraphics[align=c, height=1.3cm,width=1.3cm]{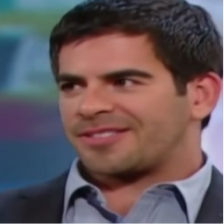}  %&\includegraphics[align=c, height=1.3cm,width=1.3cm]{image_animation_results/vox/fomm/g_0000004.png} 
  \\
  Ours & \includegraphics[align=c, height=1.3cm,width=1.3cm]{image_animation_results/vox/source.png}  &\includegraphics[align=c, height=1.3cm,width=1.3cm]{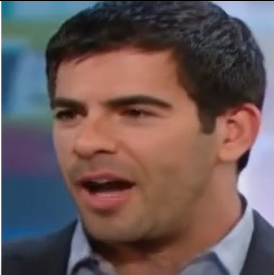} & \includegraphics[align=c, height=1.3cm,width=1.3cm]{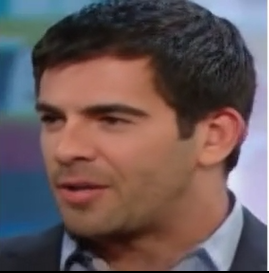} & \includegraphics[align=c, height=1.3cm,width=1.3cm]{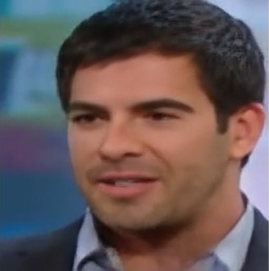} & \includegraphics[align=c, height=1.3cm,width=1.3cm]{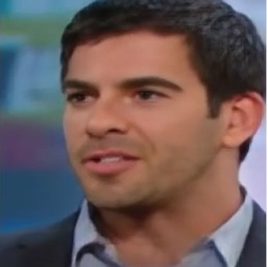} %&\includegraphics[align=c, height=1.3cm,width=1.3cm]{image_animation_results/vox/ours/g_0000004.png} 
  \\
 \midrule
 \parbox[t][][t]{1.2cm}{Tai-\\Chi}
&{\diagbox[height = 1.3cm,width=1.3cm]{\footnotesize{Source}}{\footnotesize{Driver}}}
&{\includegraphics[align=c,  height=1.3cm,width=1.3cm]{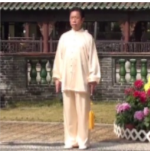}}
&{\includegraphics[align=c,  height=1.3cm,width=1.3cm]{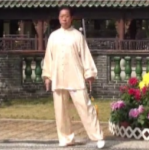}}
&{\includegraphics[align=c,  height=1.3cm,width=1.3cm]{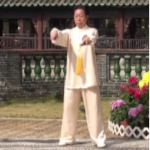}}
&{\includegraphics[align=c,  height=1.3cm,width=1.3cm]{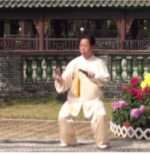}}
%&{\includegraphics[align=c,  height=1.3cm,width=1.3cm]{image_animation_results/taichi/d_0000004.png}}
\\ %\cmidrule{1-7}
% \multirow{11}{*}{\textit{\rotatebox{90}{Tai-Chi-HD}}} 
 X2Face & \includegraphics[align=c, height=1.3cm,width=1.3cm]{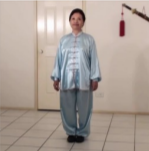} & \includegraphics[align=c, height=1.3cm,width=1.3cm]{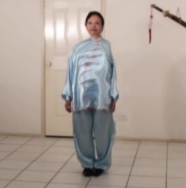} & \includegraphics[align=c, height=1.3cm,width=1.3cm]{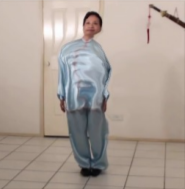} & \includegraphics[align=c, height=1.3cm,width=1.3cm]{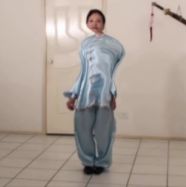} & \includegraphics[align=c, height=1.3cm,width=1.3cm]{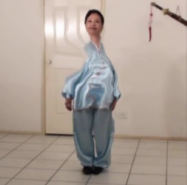} %&\includegraphics[align=c, height=1.3cm,width=1.3cm]{image_animation_results/taichi/x2face/g_0000004.png} 
 \\
  MkeyN & \includegraphics[align=c, height=1.3cm,width=1.3cm]{image_animation_results/taichi/source.png} & \includegraphics[align=c, height=1.3cm,width=1.3cm]{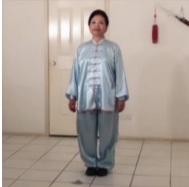} & \includegraphics[align=c, height=1.3cm,width=1.3cm]{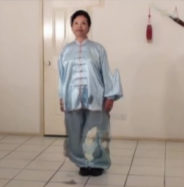} & \includegraphics[align=c, height=1.3cm,width=1.3cm]{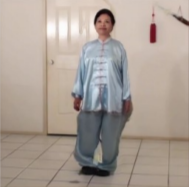} & \includegraphics[align=c, height=1.3cm,width=1.3cm]{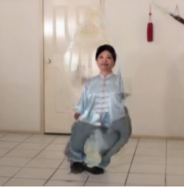} %&\includegraphics[align=c, height=1.3cm,width=1.3cm]{image_animation_results/taichi/mn/g_0000004.png} 
  \\
  FOMM & \includegraphics[align=c, height=1.3cm,width=1.3cm]{image_animation_results/taichi/source.png} & \includegraphics[align=c, height=1.3cm,width=1.3cm]{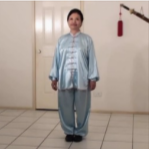} & \includegraphics[align=c, height=1.3cm,width=1.3cm]{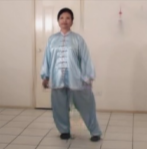} & \includegraphics[align=c, height=1.3cm,width=1.3cm]{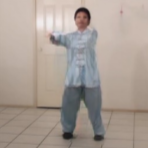} & \includegraphics[align=c, height=1.3cm,width=1.3cm]{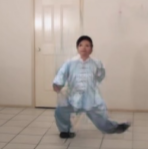} %&\includegraphics[align=c, height=1.3cm,width=1.3cm]{image_animation_results/taichi/fomm/g_0000004.png} 
  \\
  Ours & \includegraphics[align=c, height=1.3cm,width=1.3cm]{image_animation_results/taichi/source.png} & \includegraphics[align=c, height=1.3cm,width=1.3cm]{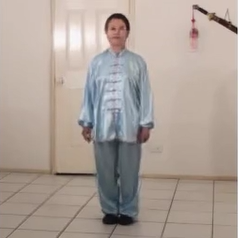} & \includegraphics[align=c, height=1.3cm,width=1.3cm]{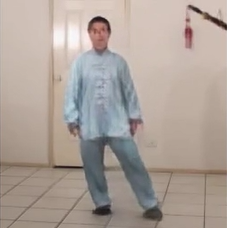} & \includegraphics[align=c, height=1.3cm,width=1.3cm]{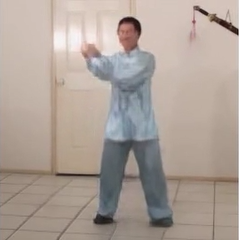} & \includegraphics[align=c, height=1.3cm,width=1.3cm]{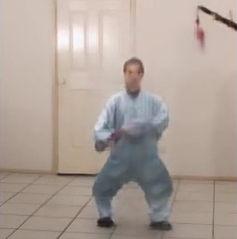} %&\includegraphics[align=c, height=1.3cm,width=1.3cm]{image_animation_results/taichi/ours/g_0000004.png} 
  \\
 \midrule
 \parbox[t][][t]{1.2cm}{BAIR}
 &{\diagbox[height = 1.3cm,width=1.3cm]{\footnotesize{Source}}{\footnotesize{Driver}}} &{\includegraphics[align=c, height=1.3cm,width=1.3cm]{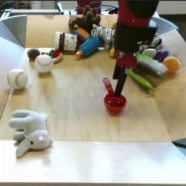}}
&{\includegraphics[align=c, height=1.3cm,width=1.3cm]{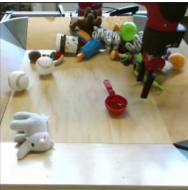}}
&{\includegraphics[align=c, height=1.3cm,width=1.3cm]{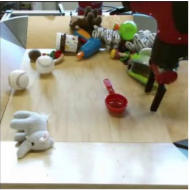}} &{\includegraphics[align=c, height=1.3cm,width=1.3cm]{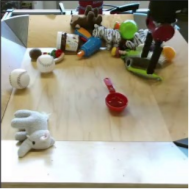}} %&{\includegraphics[align=c, height=1.3cm,width=1.3cm]{image_animation_results/robonet/d_0000004.png}}
\\ %\cmidrule{1-7}
%\multirow{11}{*}{\textit{\rotatebox{90}{BAIR}}} 
 X2Face & \includegraphics[align=c, height=1.3cm,width=1.3cm]{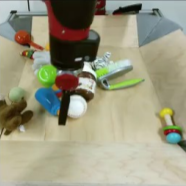} & \includegraphics[align=c, align=c,height=1.3cm,width=1.3cm]{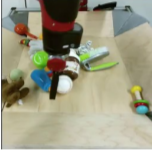} & \includegraphics[align=c, height=1.3cm,width=1.3cm]{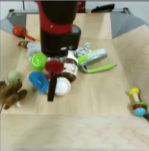} & \includegraphics[align=c, height=1.3cm,width=1.3cm]{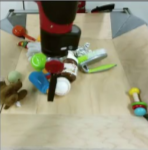} & \includegraphics[align=c, height=1.3cm,width=1.3cm]{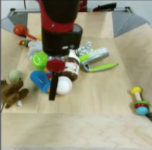} %&\includegraphics[align=c, height=1.3cm,width=1.3cm]{image_animation_results/robonet/x2face/g_0000004.png} 
 \\
 MkeyN & \includegraphics[align=c, height=1.3cm,width=1.3cm]{image_animation_results/robonet/source.png} & \includegraphics[align=c, height=1.3cm,width=1.3cm]{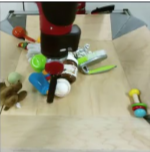} & \includegraphics[align=c, height=1.3cm,width=1.3cm]{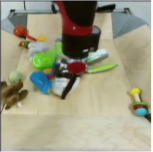} & \includegraphics[align=c, height=1.3cm,width=1.3cm]{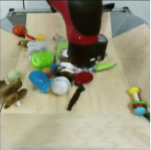} & \includegraphics[align=c, height=1.3cm,width=1.3cm]{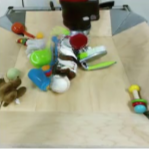} %&\includegraphics[align=c, height=1.3cm,width=1.3cm]{image_animation_results/robonet/mn/g_0000004.png} 
 \\
 FOMM & \includegraphics[align=c, height=1.3cm,width=1.3cm]{image_animation_results/robonet/source.png} & \includegraphics[align=c, height=1.3cm,width=1.3cm]{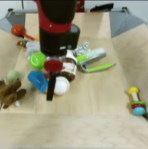} & \includegraphics[align=c, height=1.3cm,width=1.3cm]{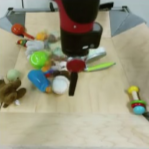} & \includegraphics[align=c, height=1.3cm,width=1.3cm]{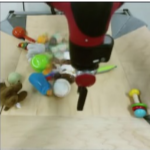} & \includegraphics[align=c, height=1.3cm,width=1.3cm]{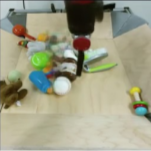} %&\includegraphics[align=c, height=1.3cm,width=1.3cm]{image_animation_results/robonet/fomm/g_0000004.png} 
 \\
 Ours & \includegraphics[align=c, height=1.3cm,width=1.3cm]{image_animation_results/robonet/source.png} & \includegraphics[align=c, height=1.3cm,width=1.3cm]{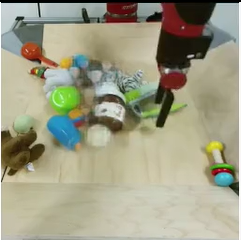} & \includegraphics[align=c, height=1.3cm,width=1.3cm]{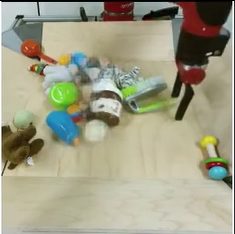} & \includegraphics[align=c, height=1.3cm,width=1.3cm]{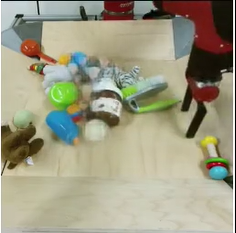} & \includegraphics[align=c, height=1.3cm,width=1.3cm]{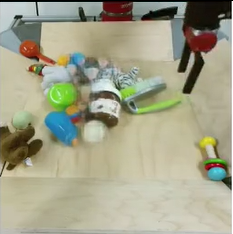} %&\includegraphics[align=c, height=1.3cm,width=1.3cm]{image_animation_results/robonet/ours/g_0000004.png}
\\
\bottomrule
\end{tabular}}
%\smallskip %endofcenter
\caption{Sample animation results on the three datasets. We use the same samples as evaluated by FOMM.}% MkeyN=Monkey-Net.}
\label{fig:qualitative_results_figure}
\vspace{-.51cm}
\end{figure}

Sample results compared to the baseline methods are shown in Fig.~\ref{fig:qualitative_results_figure}. For VoxCeleb, our method better preserves the identity of the source, and the facial expressions of the generated frames are more compatible with that of the driver. For the Tai-Chi-HD dataset, the baseline methods tend to generate infeasible poses for the fourth generated frame, while we {do not}. Unlike FOMM, we well maintain environment elements, such as the stick on the top-right of the generated frame. For the BAIR dataset, the images generated by our method are the sharpest, and it is the only method that places the generated object in the right position. {Note that the samples were selected to match those of~\cite{Siarohin_2019_CVPR}, and not by us.}

\begin{table}[t]
\centering
\resizebox{0.8\linewidth}{!}{
\begin{tabular}{@{}l@{~}cccc@{~}c@{}}
\toprule
&\multicolumn{4}{c}{\textit{Tai-Chi-HD}}
&\multicolumn{1}{c}{\textit{VoxCeleb}} \\
\cmidrule(lr){2-5}
\cmidrule(lr){6-6}
{Method} 
&\multicolumn{1}{c}{AKD$\downarrow$}
&\multicolumn{1}{c}{MKR$\downarrow$}
&\multicolumn{1}{c}{CLS$\uparrow$}
&\multicolumn{1}{c}{IOU$\uparrow$}
&\multicolumn{1}{c}{FES$\uparrow$}\\
\midrule
X2Face & 22.799 & 0.140 & 0.870 & 0.558 & 28.0\% \\
MonkeyNet & 17.308 & 0.104 & 0.852 & 0.634 & 38.2\% \\
FOMM & 10.218 & 0.044 & 0.957 & 0.864 & 48.4\% \\
Ours &  \bf 7.809 & \bf 0.020 & \bf 0.994 & \bf 0.875 & \bf 52.2\% \\
\bottomrule
\end{tabular}}
%\smallskip %endofcenter
\caption{Quantitative evaluation for image animation.}% The proposed model converges faster.}}
\label{anim_quan_table}
\end{table}

\begin{table}[t]
\centering
% \resizebox{0.8\linewidth}{!}{
\resizebox{\columnwidth}{!}{
\begin{tabular}{@{}l@{~}c@{~}c@{~}c@{~}c@{~}c@{~}c@{~}c@{~}c@{}}
\toprule
{Model}
&{X2Face}
&{MN}
&{FOMM}
&{no\_pert}
&{no\_ref}
&{no\_id}
&{low\_res}
&{Ours}
\\ \midrule
\textit{OpenFace \cite{amos2016openface}} & 0.512 &	0.544 &	0.620 &	0.625 &	0.487 &	0.522 &	0.632 &	\bf 0.642
\\
\textit{DeepFace \cite{6909616}} & 0.528 &	0.580 &	0.646 &	0.648 &	0.515 &	0.546 &	0.658 &	\bf 0.676
\\
\textit{DeepID \cite{2014arXiv1406.4773S}} & 0.799 &	0.827 &	0.953 &	0.917 &	0.756 &	0.786 &	0.948 &	\bf 0.963
\\
\bottomrule
\end{tabular}}
%\smallskip %endofcenter
\caption{CSIM for VoxCeleb, including the ablation models.}
\label{CSIM_anim_table}
\end{table}

\begin{table*}
\centering
\resizebox{0.7\linewidth}{!}{
\begin{tabular}{lccc}
\toprule
{Method} 
&\multicolumn{1}{c}{AKD $\downarrow$}
&\multicolumn{1}{c}{MKR$\downarrow$} 
&\multicolumn{1}{c}{FES $\uparrow$}\\
\midrule
FOMM & 10.218\textbackslash 8.629\textbackslash 9.958\textbackslash 12.364 & 0.044\textbackslash 0.042\textbackslash 0.042\textbackslash 0.049 & 48.4\% \textbackslash 48.9\% \textbackslash 50.7\% \textbackslash 45.6\% \\
Ours &  \bf 7.809\textbackslash 6.431\textbackslash 7.433\textbackslash 8.909 & \bf 0.020\textbackslash 0.017\textbackslash 0.020\textbackslash 0.025 & \bf 52.2\% \textbackslash 54.3\% \textbackslash 53\% \textbackslash \bf 49.3\% \\
\bottomrule
\end{tabular}}
%\smallskip %endofcenter
\caption{AKD and MKR for Tai-Chi-HD. FES for VoxCeleb. All are reported for the Full\textbackslash \nth{1} \textbackslash \nth{2} \textbackslash \nth{3} sets.}
\label{groups_tab}
\end{table*}

\begin{figure}[t]
\centering
\resizebox{0.9\linewidth}{!}
{\begin{tabular}{@{}c@{~}c@{}c@{}c@{}c@{}c@{}c@{}c@{}c@{}c@{}c@{}c@{}c@{}}
\multirow{4}{*}{\textit{\rotatebox{90}{Tai-Chi-HD}}} &
\includegraphics[height=\size1 cm,width=\size1 cm]{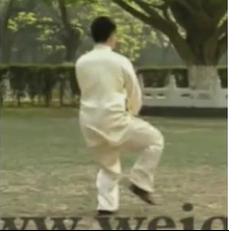} & \includegraphics[height=\size1 cm,width=\size1 cm]{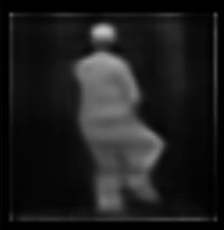} &
\includegraphics[height=\size1 cm,width=\size1 cm]{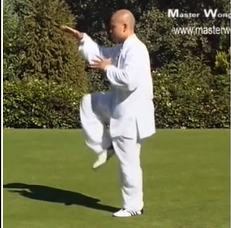} & \includegraphics[height=\size1 cm,width=\size1 cm]{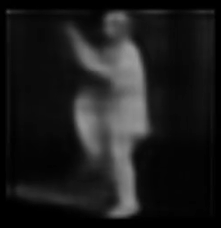} & \includegraphics[height=\size1 cm,width=\size1 cm]{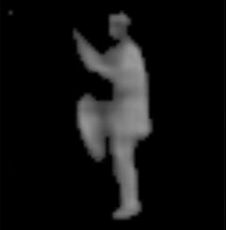} & \includegraphics[height=\size1 cm,width=\size1 cm]{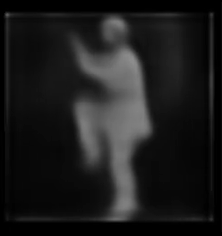}\\ 
& $\vs$ & $\vm_s$ & $\vd$ & $\vm_d$ & $\vm_{dp}$ & $\vm_{dr}$  \\
 &
\includegraphics[height=\size1 cm,width=\size1 cm]{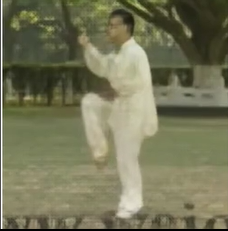} & \includegraphics[height=\size1 cm,width=\size1 cm]{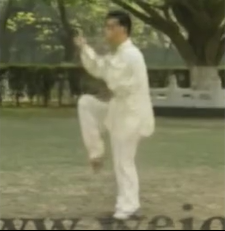} & \includegraphics[height=\size1 cm,width=\size1 cm]{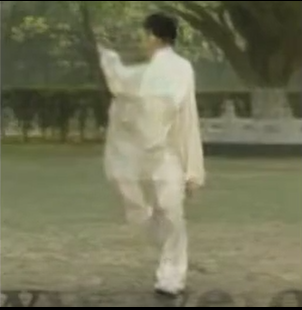} &
\includegraphics[height=\size1 cm,width=\size1 cm]{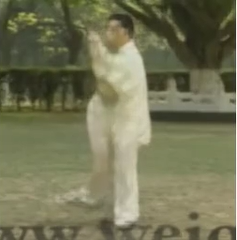} &
\includegraphics[height=\size1 cm,width=\size1 cm]{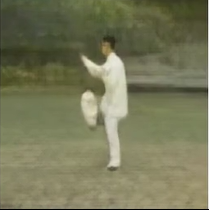} &
\includegraphics[height=\size1 cm,width=\size1 cm]{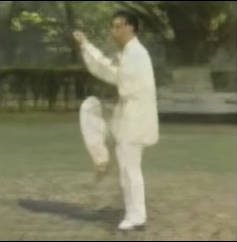} \\
 & $\vc$ & $\vf$ & FOMM & {no\_pert} & {no\_ref} & {no\_id}  \\
\multirow{4}{*}{\textit{\rotatebox{90}{VoxCeleb}}} &
\includegraphics[height=\size1 cm,width=\size1 cm]{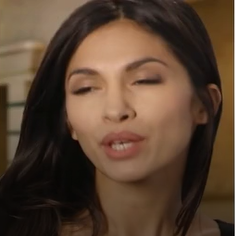} & \includegraphics[height=\size1 cm,width=\size1 cm]{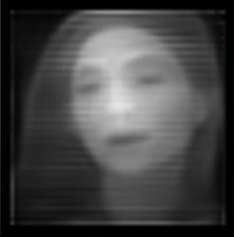} &
\includegraphics[height=\size1 cm,width=\size1 cm]{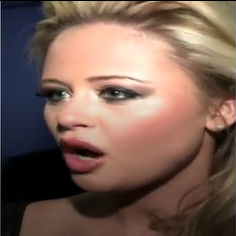} & \includegraphics[height=\size1 cm,width=\size1 cm]{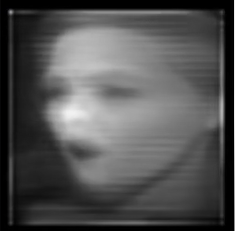} & \includegraphics[height=\size1 cm,width=\size1 cm]{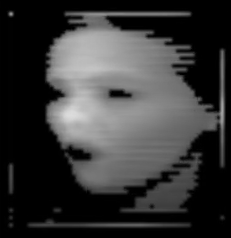} & \includegraphics[height=\size1 cm,width=\size1 cm]{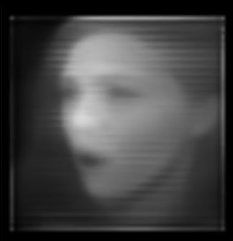} \\
& $\vs$ & $\vm_s$ & $\vd$ & $\vm_d$ & $\vm_{dp}$ & $\vm_{dr}$  \\
%\textit{\rotatebox{90}{~~~VoxCeleb}} &
&\includegraphics[height=\size1 cm,width=\size1 cm]{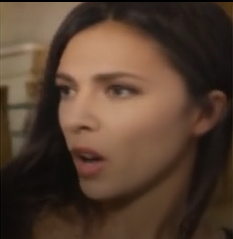} & \includegraphics[height=\size1 cm,width=\size1 cm]{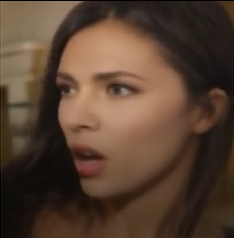} & \includegraphics[height=\size1 cm,width=\size1 cm]{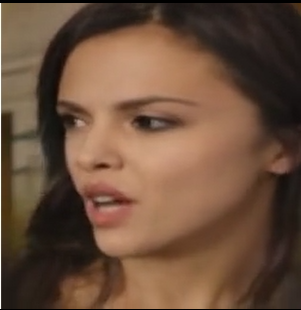} &
\includegraphics[height=\size1 cm,width=\size1 cm]{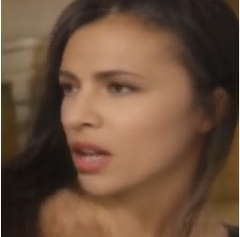} &
\includegraphics[height=\size1 cm,width=\size1 cm]{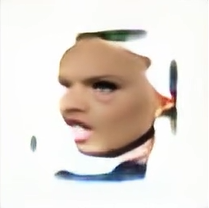} &
\includegraphics[height=\size1 cm,width=\size1 cm]{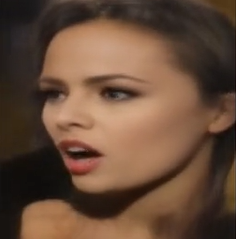} \\
 & $\vc$ & $\vf$ & FOMM & {no\_pert} & {no\_ref} & {no\_id}  \\
\multirow{4}{*}{\textit{\rotatebox{90}{BAIR}}} &
\includegraphics[height=\size1 cm,width=\size1 cm]{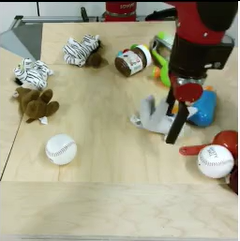} & \includegraphics[height=\size1 cm,width=\size1 cm]{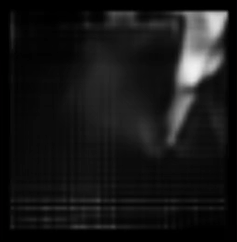} &
\includegraphics[height=\size1 cm,width=\size1 cm]{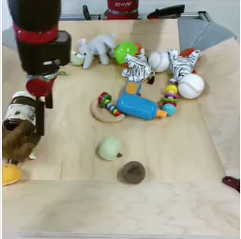} & \includegraphics[height=\size1 cm,width=\size1 cm]{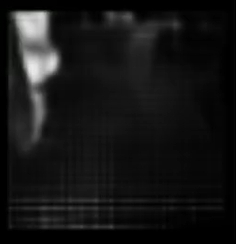} & \includegraphics[height=\size1 cm,width=\size1 cm]{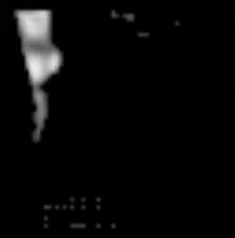} & \includegraphics[height=\size1 cm,width=\size1 cm]{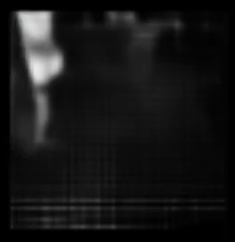} \\
& $\vs$ & $\vm_s$ & $\vd$ & $\vm_d$ & $\vm_{dp}$ & $\vm_{dr}$  \\
%\textit{\rotatebox{90}{~~~~~BAIR}} 
&
\includegraphics[height=\size1 cm,width=\size1 cm]{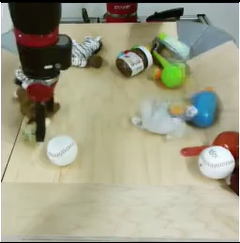} & \includegraphics[height=\size1 cm,width=\size1 cm]{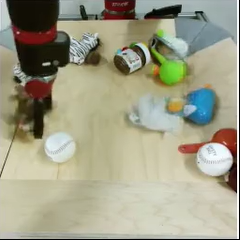} & \includegraphics[height=\size1 cm,width=\size1 cm]{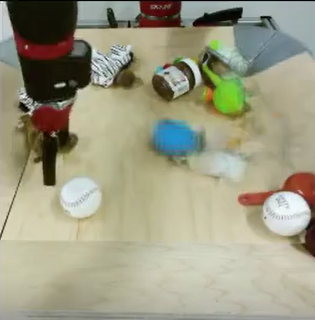} &
\includegraphics[height=\size1 cm,width=\size1 cm]{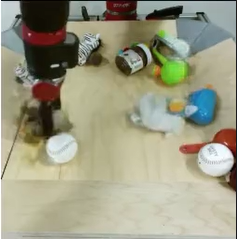} &
\includegraphics[height=\size1 cm,width=\size1 cm]{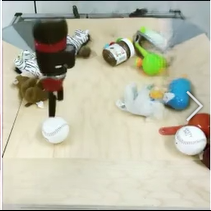} &
\includegraphics[height=\size1 cm,width=\size1 cm]{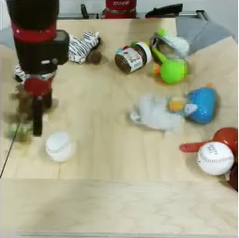} \\
 & $\vc$ & $\vf$ & FOMM & {no\_pert} & {no\_ref} & {no\_id}  \\
\end{tabular}}
%\smallskip %endofcenter
\caption{Intermediate results generated by our method. The generated frame $\vf$ is compared to FOMM and ablation models. From left to right: the source frame $\vs$, source mask $\vm_s$, the driving frame $\vd$, the driving mask $\vm_d$, the perturbed driving mask $\vm_{dp}$, the refined driving mask $\vm_{dr}$, the low-res prediction $\vc$, the high-res prediction $\vf$, FOMM's result, and the ablations: {no\_pert} drops $\tP_\text{test}$, {no\_ref} omits the mask refinement $\tR$, and {no\_id} omits both.}
\label{fig:generated_sequences_figure}
\end{figure}

\smallskip

\noindent{\bf Ablation\quad}
\label{ablation_sec}
The main challenge is to replace the identity on the driver's mask with that of the source while keeping on the driver's pose. We do that in two steps: (i) the driver's identity is interrupted by applying $\tP_{\text{test}}$, (ii) the refinement network $\tR$ acts to replace the driver's identity with that of the source. To evaluate the roles of $\tP_{\text{test}}$ and $\tR$, we evaluate three partial methods: {no\_pert}, {no\_ref}, and {no\_id}, where the first, second, or both steps are removed, respectively.

Ablation and intermediate results generated by our pipeline are shown in Fig.~\ref{fig:generated_sequences_figure}. As can be seen, the generated masks $\vm_s$ and $\vm_d$ capture very accurately the object's pose and shape, and the mask refinement network $\tR$ successfully apply the source's identity to the driver's mask. Comparing the generated frame $\vf$ to that of FOMM, we notice that for the Tai-Chi-HD dataset, the pose of the generated body using our method is much more compatible with that of the driver's, where FOMM's model generates a distorted body. 

For VoxCeleb, using our method, the identity of the source is better preserved, as it also reveals a small portion of the teeth, as the driver does. For the BAIR dataset, unlike FOMM, our method was able to inpaint the occluded surface, including the white and blue items on the right of the generated frame. Examining the generated frames of the ablation models shows that both steps, identity-perturbation and mask refinement, are critical. The frames generated by {no\_pert} and no\_id have significant traces of the driver's identity. This is especially clear for VoxCeleb, on the forehead area of {no\_pert} and the general appearance for {no\_id}. Similarly, for Tai-Chi-HD, the frame generated by {no\_ref} contains traces from the driver's environment, and for the other datasets, it generates distorted results.

Next, we evaluated the following ablation models. \textit{no\_color\_aug}, where the color augmentation is eliminated at train time. \textit{h\_update\_l}, where the high-res generator $\tH$ keeps updating the weights of the low-res generator $\tL$. \textit{h\_update\_m}, where the high-res generator $\tH$ keeps updating the weights of the mask generator $\tM$. The ablation models were trained on the VoxCeleb dataset and evaluated on the video reconstruction task. Results are presented in Tab.~\ref{ablations_results_table}. As can be seen, using the color augmentation and limiting the task of the high-res generator $\tH$ for adding fine details, helps the model to converge faster.

Next, we analyze the importance of the suggested modules for identity preservation, using the CSIM between source and generated frames. The results are shown in Tab.~\ref{CSIM_anim_table}. As can be seen, removing the refinement step ({no\_ref}, {no\_id}) dramatically degrades the CSIM score, and applying $\tP_{\text{test}}$ helps $\tR$ to better inject the source’s identity. It is also can be seen that low\_res results in a lower CSIM score, which verifies the effectiveness of $\tH$. 

In Fig.~\ref{fig:low_res_vs_high_res} we show an example for the visual improvement of $\vf$ over $\vc$. The environment in both examples and the face of the man in the left example are much sharper in $\vf$. In addition, we show an example where the generated masks reflects very well whether the subject is facing back or not.

\begin{table}[t]
\centering
\resizebox{0.65\linewidth}{!}{
\begin{tabular}{lccc}
\toprule
{Method} 
&\multicolumn{1}{c}{L1}
&\multicolumn{1}{c}{AKD}
&\multicolumn{1}{c}{AED} \\
\midrule
{No\_color\_aug} & 0.045& 1.863& 0.159 \\
{H\_update\_m} & 0.041& 1.829& 0.161 \\
{H\_update\_l}  & 0.039& 1.412& 0.142 \\
{Full method} & \bf 0.034& \bf 1.329& \bf 0.130 \\
\bottomrule
\end{tabular}}
%\smallskip %endofcenter
\caption{Ablation analysis on the reconstruction task for VoxCeleb.}% The proposed model converges faster.}}
\label{ablations_results_table}
\end{table}

\begin{table}[b]
\centering
\resizebox{0.9\linewidth}{!}
{\begin{tabular}{@{}c@{}c@{}c@{}c@{}|@{}c@{}c@{}}
\includegraphics[height=\size1 cm,width=\size1 cm]{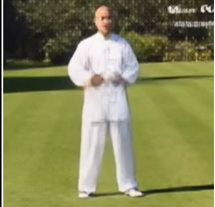} & \includegraphics[height=\size1 cm,width=\size1 cm]{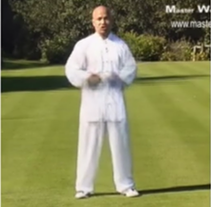} &
\includegraphics[height=\size1 cm,width=\size1 cm]{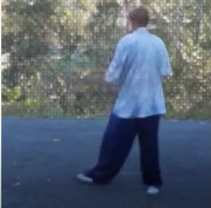} &
\includegraphics[height=\size1 cm,width=\size1 cm]{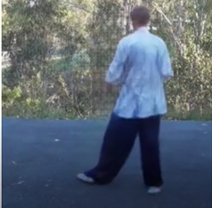}&\includegraphics[height=\size1 cm,width=\size1 cm]{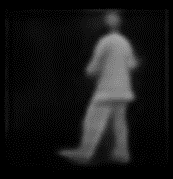}& \includegraphics[height=\size1 cm,width=\size1 cm]{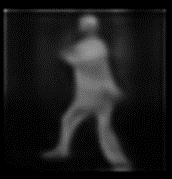} \\
$\vc$ & $\vf$ & $\vc$ & $\vf$  & back & front\\
\end{tabular}}
%\smallskip %endofcenter
\captionof{figure}{(left) $\vf$ is sharper than $\vc$. (right) back \& front masks.}
\label{fig:low_res_vs_high_res}
\end{table}

\smallskip

\noindent{\bf User study\quad} To further qualitatively evaluate our method and compare it with existing work, we presented volunteers with a source image, a driving video, and four randomly ordered generated videos, one for each baseline method. They were asked to (i) select the most realistic animation of the source image, and (ii) select the video with the highest fidelity to the driver video. For each of the $n=25$ participants, we repeated the experiment three times, each time using a different dataset and a random test sample. 

The results, see Tab.~\ref{userstudy_results_table}, are highly consistent with the quantitative results, and indicate that the quality and the animation of the videos our method generated, contain fewer artifacts and are better synchronized with the driver videos.
In addition, it can be seen that the refinement network $\tR$ is the most important module for quality and for motion, and that the perturbation operator $\tP_\text{test}$ is much more needed in Tai-Chi-HD and VoxCeleb.

\begin{table}[t]
% \centering
% \begin{tabular}{@{}l@{~~}lcccc@{}}
% \toprule
%  & 
% {Dataset}
% &{X2Face}
% &{MN}
% &{FOMM}
% &{Ours}
% \\ \midrule
% \multirow{3}{*}{\textit{\rotatebox{90}{Quality}}} & \textit{Tai-Chi-HD} & 0\% & 4\% & 16\% & 80\% 
% \\
% &\textit{VoxCeleb} & 0\% & 8\% & 16\% & 76\% 
% \\
% &\textit{BAIR} & 0\% & 4\% & 24\% & 72\% 
% \\
%  \midrule
% \multirow{3}{*}{\textit{\rotatebox{90}{Motion}}} & \textit{Tai-Chi-HD} & 0\% & 4\% & 8\% & 88\% 
% \\
% &\textit{VoxCeleb} & 0\% & 0\% & 12\% & 88\% 
% \\
% &\textit{BAIR} & 4\% & 4\% & 16\% & 76\% 
% \\
% \bottomrule
% \end{tabular}
% %\smallskip %endofcenter
% \caption{The ratio of best videos selected for each method, based on quality and motion fidelity. MN=Monkey-Net. }
% \label{userstudy_results_table}
\centering
% \resizebox{0.8\linewidth}{!}{
\resizebox{\columnwidth}{!}{
\begin{tabular}{l@{~}c@{~}c@{~}c@{~}c@{~}c@{~}c@{~}c@{}}
\toprule
{Dataset}
&{X2Face}
&{MN}
&{FOMM}
&{no\_pert}
&{no\_ref}
&{no\_id}
&{Ours}
\\ \midrule
\textit{Tai-Chi} & (0\%,0\%) & (4\%,4\%) & (16\%,8\%) & (2\%,2\%) & (0\%,0\%) & (0\%,0\%) & (\bf 78\%, \bf 86\%)
\\
\textit{VoxCeleb} & (0\%,0\%) & (6\%,4\%) & (10\%,10\%) & (12\%,10\%) & (0\%,0\%) & (0\%,0\%) & \bf (72\%,\bf 76\%)
\\
\textit{BAIR} & (0\%,4\%) & (6\%,6\%) & (14\%,8\%) & (20\%,16\%) & (0\%,0\%) & (0\%,0\%) & (\bf 60\%, \bf 66\%)
\\
\bottomrule
\end{tabular}}
%\smallskip %endofcenter
\caption{The ratio (Quality, Motion-fidelity) of best videos selected for each method, including ablations. MN=Monkey-Net.}
\label{userstudy_results_table}
\end{table}

\smallskip

\noindent{\bf Limitations\quad}
While better than the baselines, there are artifacts and identity loss for extreme changes in pose and shape. Additionally, since the perturbation operator sets to zero mask's pixels that are smaller than a threshold $\rho$, some pose information may be lost. Other failure cases are ambiguities on generated masks, e.g. for Tai-Chi-HD, when the hands are overlapped, the generator may struggle to understand which one is on top. This limitation also exists for the baseline methods including key-points methods. 

% Other advanced methods may be used for the decision on which pixels should be removed, and we leave it for a future work.

As a video generation method, considerations should be made towards the possible use of the generated output in a harmful way. For example, the generated videos of talking heads can be used as part of a system for manipulating speech content. Our hope is that studying such methods in an open way would enable the mitigation of such risks through better detection methods and by raising awareness.

\section{Conclusions}

A novel method for conditionally reanimating a frame is presented. It utilizes a masking mechanism for encoding pose information. Our method is able to effectively extract both the source and the driving masks, while accurately capturing the shape and foreground/background separation, and recovering an identity-free pose representation of the driver. Our results outperform the state of the art by a sizable margin on the available benchmarks.

% \appendix
% \section{Architecture}
% \label{arch_section}
% Following \cite{siarohin2019first}, the mask generator $\tM$, the mask refinement network $\tR$ and the high-res generator $\tH$ have the same encoder-decoder architecture, followed by a $conv_{7\stimes7}$ layer and a $sigmoid$ activation. The encoder (decoder) consists of five encoding (decoding) blocks, where each encoding block is a sequence of $conv_{3\stimes3} - relu - batch\_norm - avg\_pool_{2\stimes2}$, and each decoding block is a sequence of $up\_sample_{2\stimes2} - conv_{3\stimes3} - batch\_norm - relu$. Only for the high-res generator $\tH$ we add skip connections from each of the encoding layers, to its corresponding decoding layer, to form a U-Net architecture.

% The encoder of the low-res generator $\tL$ consists of $conv_{7\stimes7} - batch\_norm - relu$ , followed by six residual blocks, each block consisting of $batch\_norm - relu - conv_{3\stimes3} - batch\_norm - relu - conv_{3\stimes3}$. 
% The decoder consists of two blocks, each is a sequence of $up\_sample_{2\stimes2} - conv_{3\stimes3} - batch\_norm - relu$.
% The decoder is followed by a $conv_{7\stimes7}$ layer and a $sigmoid$ activation.

\section*{Acknowledgments}
This project has received funding from the European Research Council (ERC) under the European Union's Horizon 2020 research and innovation programme (grant ERC CoG 725974). The contribution of the first author is part of a PhD thesis research conducted at Tel Aviv University.

%%%%%%%%% REFERENCES
{\small
\bibliographystyle{ieee_fullname}
\bibliography{main}
}
% \clearpage
\appendix 

\section*{Appendices}
In the following appendices we present the architecture of the model (\cref{sec:arch}), discuss about the effect of using a wider bottleneck for the baselines (\cref{sec:bottleneck}), and provide additional qualitative results (\cref{sec:qres}).

\section{Architecture}
\label{sec:arch}
Following \cite{siarohin2019first}, the mask generator $\tM$, the mask refinement network $\tR$ and the high-res generator $\tH$ have the same encoder-decoder architecture, followed by a $conv_{7\stimes7}$ layer and a $sigmoid$ activation. The encoder (decoder) consists of five encoding (decoding) blocks, where each encoding block is a sequence of $conv_{3\stimes3} - relu - batch\_norm - avg\_pool_{2\stimes2}$, and each decoding block is a sequence of $up\_sample_{2\stimes2} - conv_{3\stimes3} - batch\_norm - relu$. Only for the high-res generator $\tH$ we add skip connections from each of the encoding layers to its corresponding decoding layer, to form a U-Net architecture.

The encoder of the low-res generator $\tL$ consists of $conv_{7\stimes7} - batch\_norm - relu$ , followed by six residual blocks, each block consists of $batch\_norm - relu - conv_{3\stimes3} - batch\_norm - relu - conv_{3\stimes3}$. 
The decoder consists of two blocks, each is a sequence of $up\_sample_{2\stimes2} - conv_{3\stimes3} - batch\_norm - relu$.
The decoder is followed by a $conv_{7\stimes7}$ layer and a $sigmoid$ activation.

\section{Bottleneck Size}
\label{sec:bottleneck}
In order to verify that the improvement over the baselines is not due to their smaller bottleneck size, we re-trained FOMM~\cite{siarohin2019first} and MonkeyNet~\cite{Siarohin_2019_CVPR} on all three datasets, VOXCeleb~\cite{Nagrani17}, Tai-Chi-HD~\cite{siarohin2019first}, and BAIR~\cite{2017arXiv171005268E}, using a wider bottleneck, and evaluated the video reconstruction task. We used $365$ key-points for FOMM, which are equivalent to $2190$ floating-point numbers, and $440$ key-points for MonkeyNet, which are equivalent to $2200$ floating-point numbers. As reported in Tab.~\ref{wider_bottleneck_table}, we saw no improvement.

\begin{table}
\centering
\resizebox{1\linewidth}{!}{
\begin{tabular}{@{}l@{~}c@{~}c@{~}c@{~}c@{~}c@{~}c@{~}c@{~}c@{}}
\toprule
&\multicolumn{4}{c}{\textit{Tai-Chi-HD}} 
&\multicolumn{3}{c}{\textit{VoxCeleb}} 
&\multicolumn{1}{c}{\textit{BAIR}}\\
\cmidrule(lr){2-5}
\cmidrule(lr){6-8}
\cmidrule(lr){9-9}
{Method} 
&\multicolumn{1}{c}{L1}
&\multicolumn{1}{c}{AKD}
&\multicolumn{1}{c}{MKR}
&\multicolumn{1}{c}{AED} 
&\multicolumn{1}{c}{L1}
&\multicolumn{1}{c}{AKD}
&\multicolumn{1}{c}{AED} 
&\multicolumn{1}{c}{L1}\\
\midrule
MN~\cite{Siarohin_2019_CVPR}  & \multicolumn{8}{c}{-- ~The wider bottleneck model diverged~ --}\\
FOMM~\cite{siarohin2019first} & 0.068 & 8.561 & 0.043 & 0.196 & 0.050 & 1.525 & 0.165 & 0.028 \\
Ours &  \bf 0.047 & \bf 4.239 & \bf 0.015 & \bf 0.147 & \bf 0.034 & \bf 1.329 & \bf 0.130 & \bf 0.021\\
% \midrule
% Improvement &  37\% & 61\% & 10\% & 32\% & 54\% & 88\% & 54\% & 31\%\\
\bottomrule
\end{tabular}
}
%%\smallskip %endofcenter
\smallskip
\caption{{Video reconstruction using a wider bottleneck for baselines. MN=MonkeyNet}}
\label{wider_bottleneck_table}
\end{table}

\section{Additional Qualitative Results}
\label{sec:qres}
In Fig.~\ref{fig:vox_examples}, Fig.~\ref{fig:tai_examples} and Fig.~\ref{fig:bair_examples} we added final and intermediate results generated by our method for the VoxCeleb, Tai-Chi-HD and BAIR datasets, compared to the SOTA methods. The full videos are available at https://github.com/itsyoavshalev/Image-Animation-with-Perturbed-Masks.

% {\small
% \bibliographystyle{ieee_fullname}
% \bibliography{egbib}
% }

\clearpage

\begin{figure*}[t]
	\centering
	\caption{Final and intermediate results generated by our method for VoxCeleb, compared to the SOTA methods.}
    \includegraphics[width=0.95\linewidth]{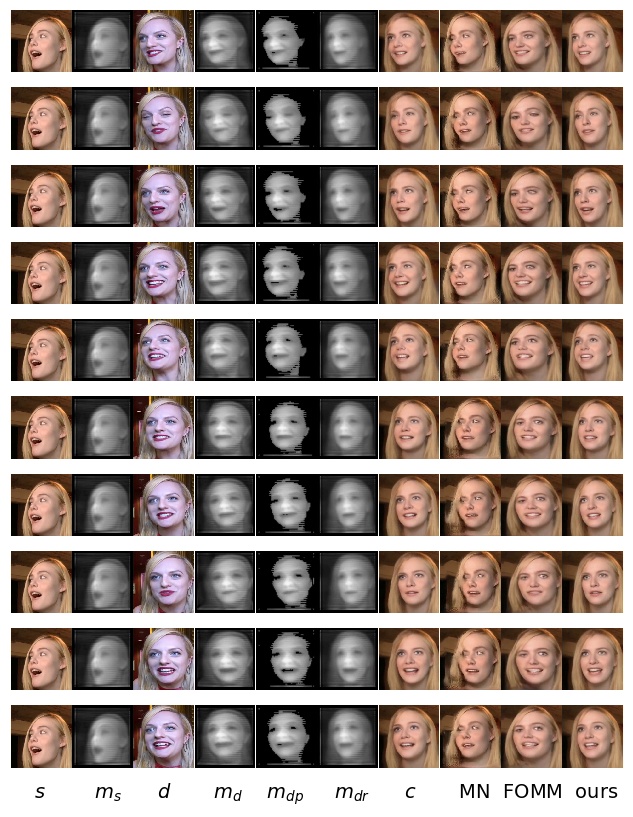}
        \label{fig:vox_examples}
\end{figure*}
\begin{figure*}[h!]
	\centering
    \ContinuedFloat
    \includegraphics[width=0.95\linewidth]{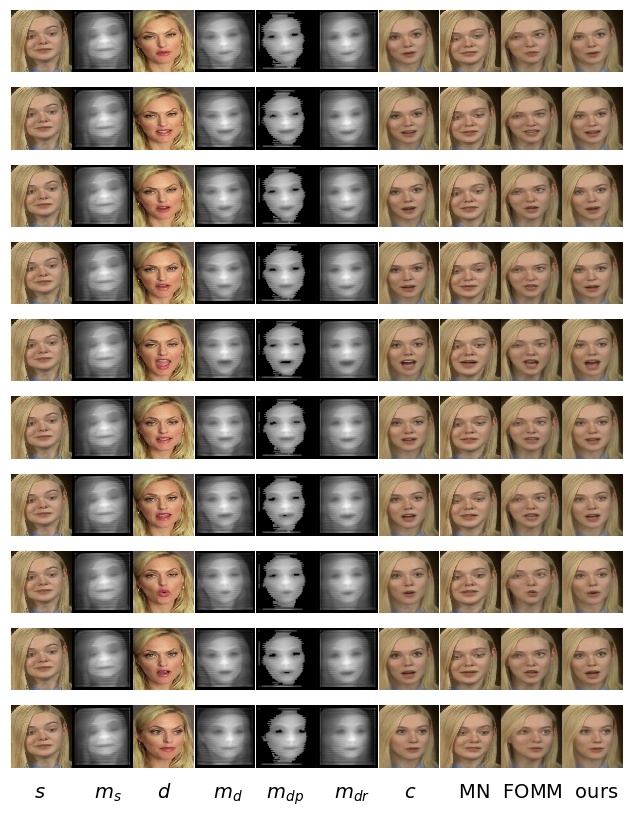}
\end{figure*}
\begin{figure*}[h!]
	\centering
    \ContinuedFloat
    \includegraphics[width=0.95\linewidth]{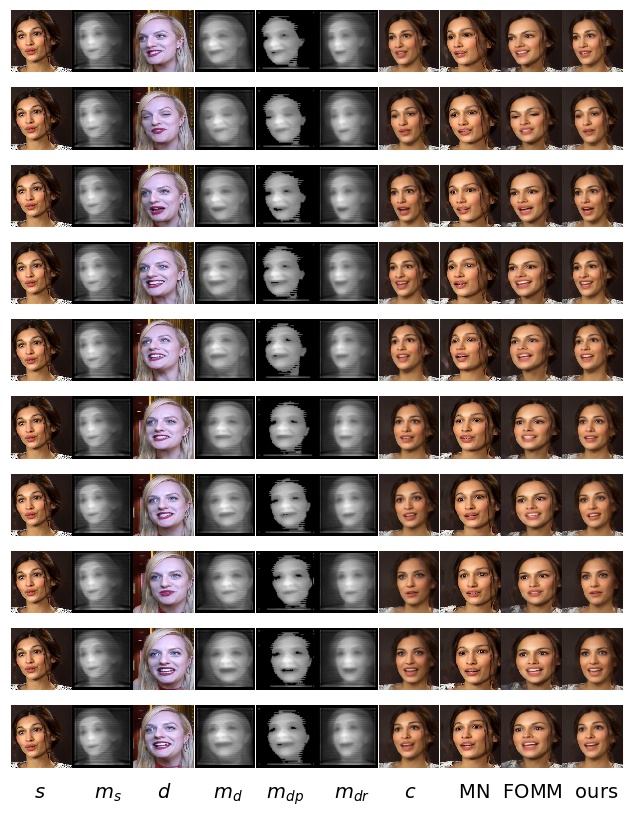}
\end{figure*}
\begin{figure*}[h!]
	\centering
    \ContinuedFloat
    \includegraphics[width=0.95\linewidth]{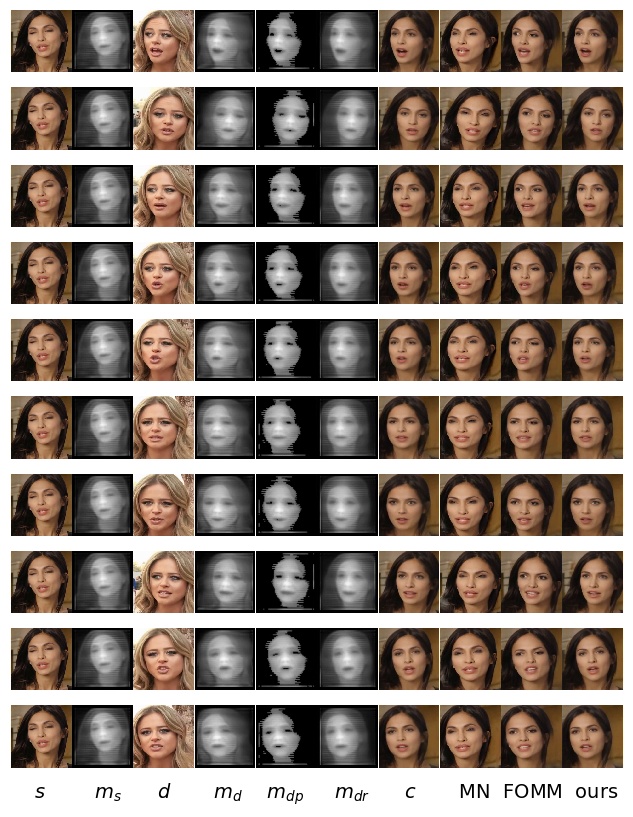}
\end{figure*}
\begin{figure*}[h!]
	\centering
    \ContinuedFloat
    \includegraphics[width=0.95\linewidth]{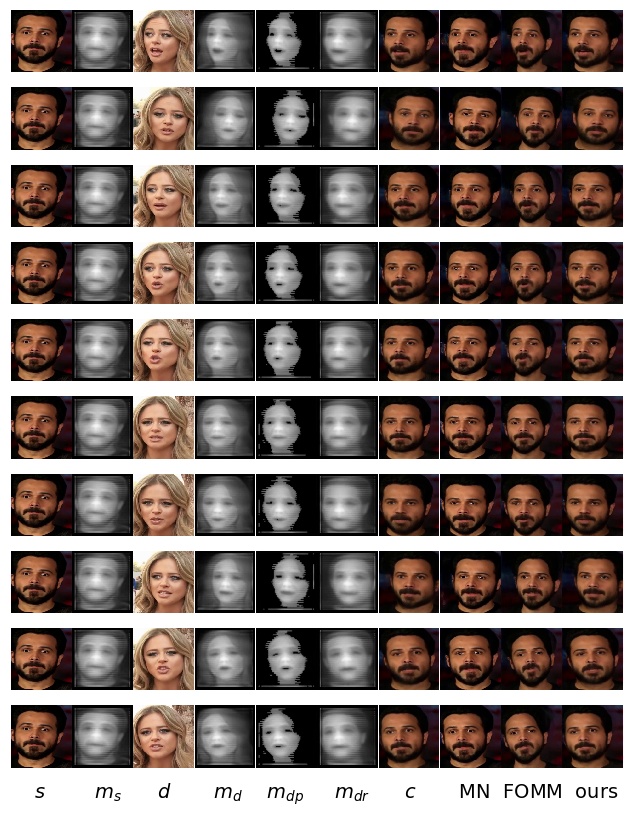}
\end{figure*}
\begin{figure*}[h!]
	\centering
    \ContinuedFloat
    \includegraphics[width=0.95\linewidth]{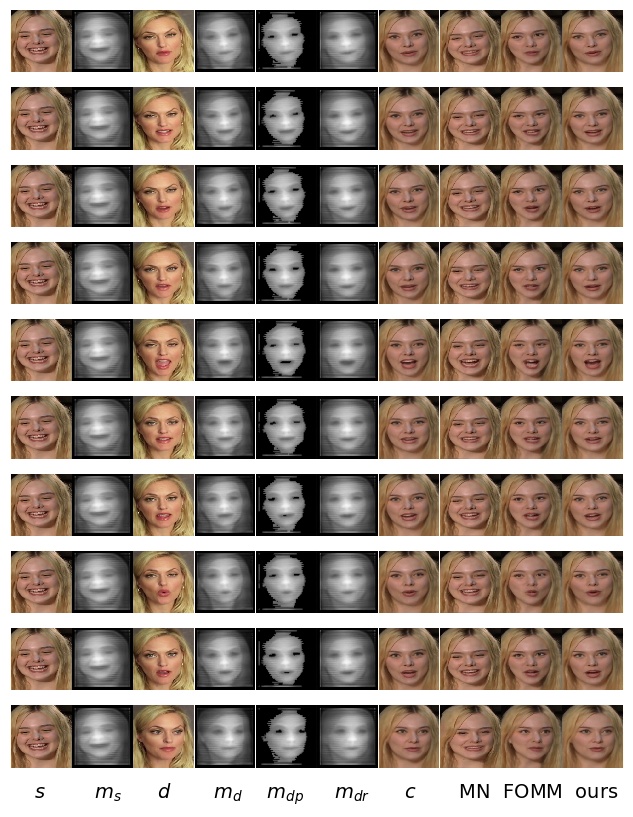}
\end{figure*}
\begin{figure*}[h!]
	\centering
    \ContinuedFloat
    \includegraphics[width=0.95\linewidth]{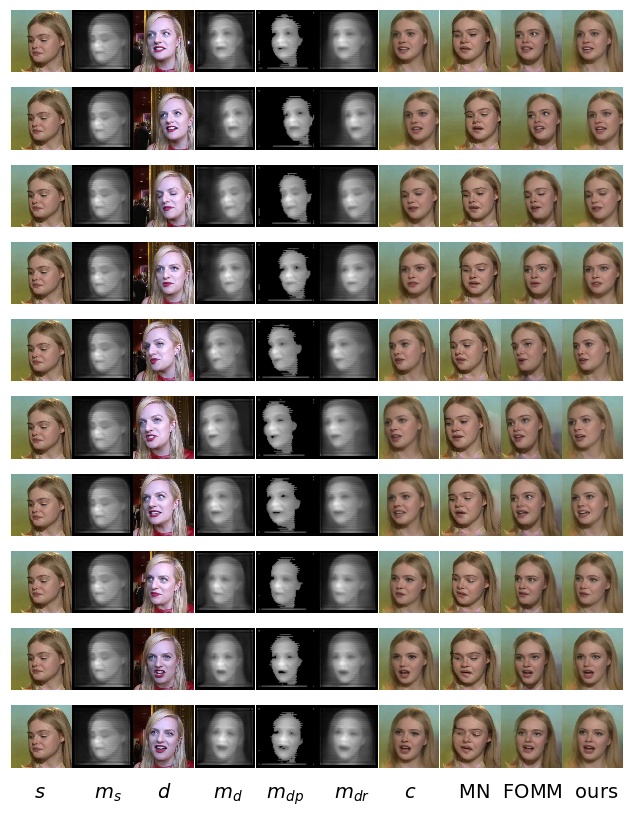}
\end{figure*}
\begin{figure*}[h!]
	\centering
    \ContinuedFloat
    \includegraphics[width=0.95\linewidth]{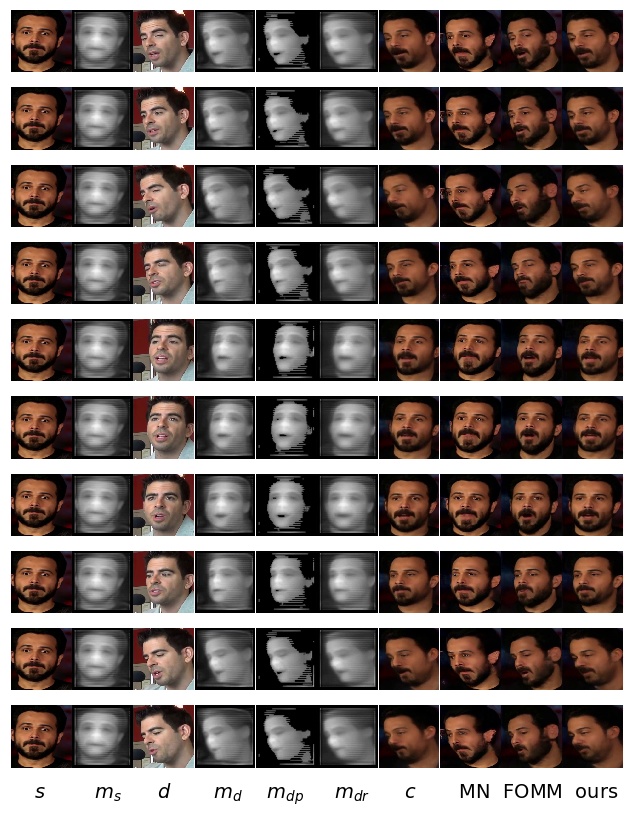}
\end{figure*}

\begin{figure*}[t]
	\centering
	\caption{Final and intermediate results generated by our method for Tai-Chi-HD, compared to the SOTA methods.}
    \includegraphics[width=0.95\linewidth]{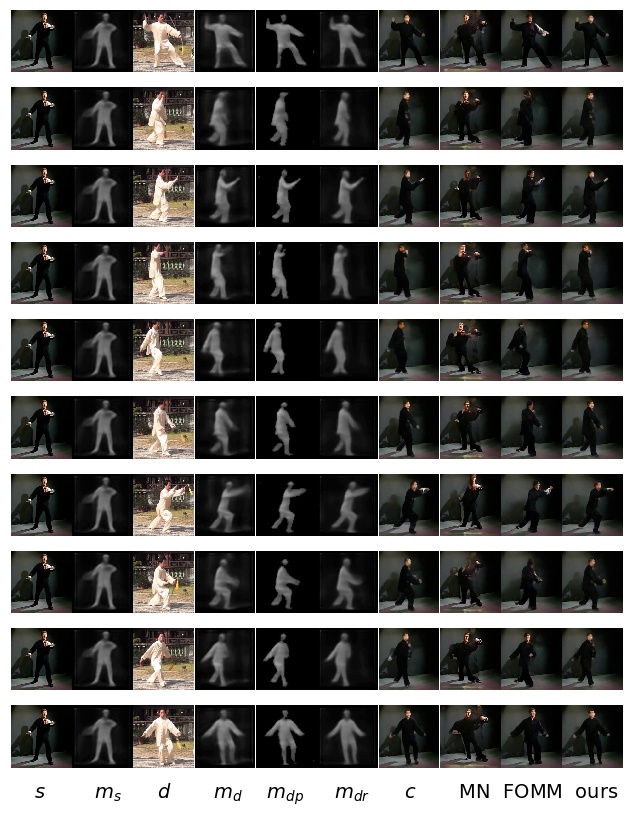}
        \label{fig:tai_examples}
\end{figure*}
\begin{figure*}[h!]
	\centering
    \ContinuedFloat
    \includegraphics[width=0.95\linewidth]{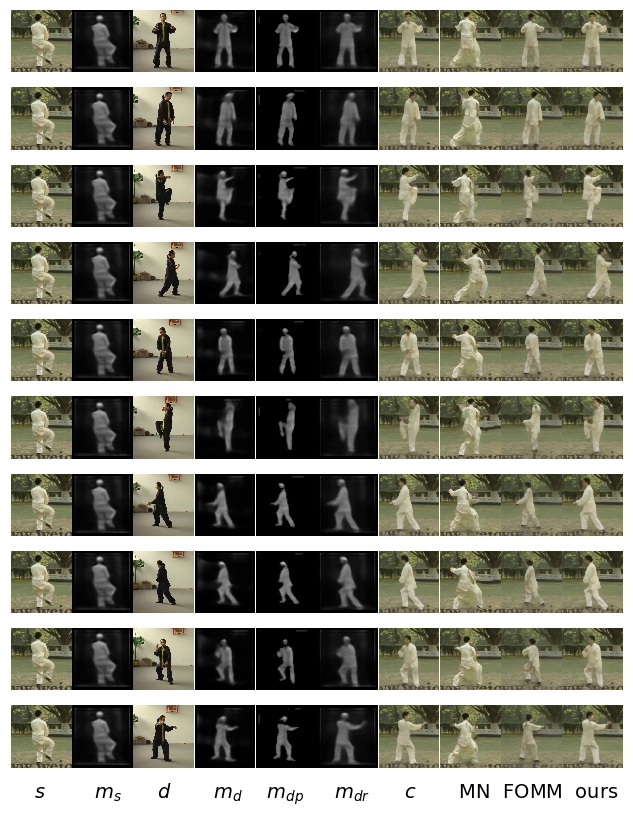}
\end{figure*}
\begin{figure*}[h!]
	\centering
    \ContinuedFloat
    \includegraphics[width=0.95\linewidth]{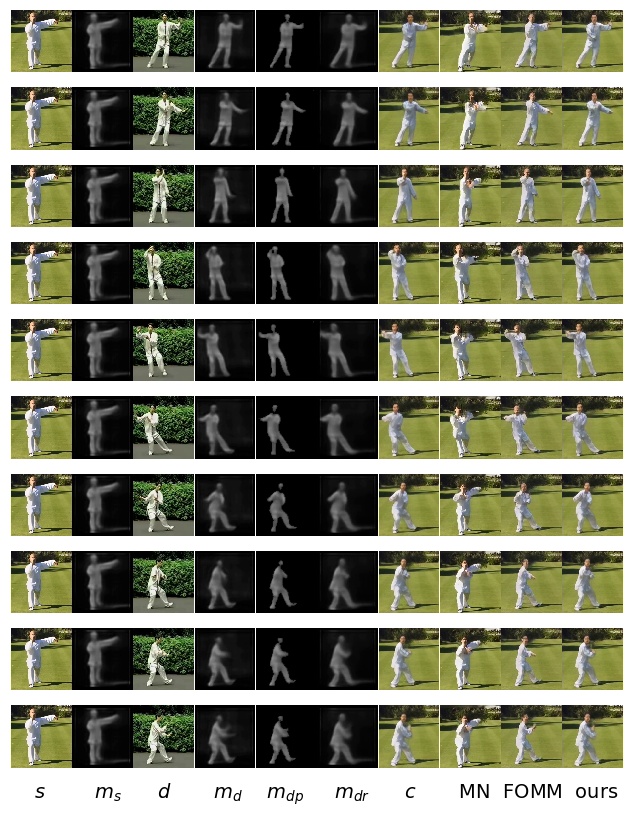}
\end{figure*}
\begin{figure*}[h!]
	\centering
    \ContinuedFloat
    \includegraphics[width=0.95\linewidth]{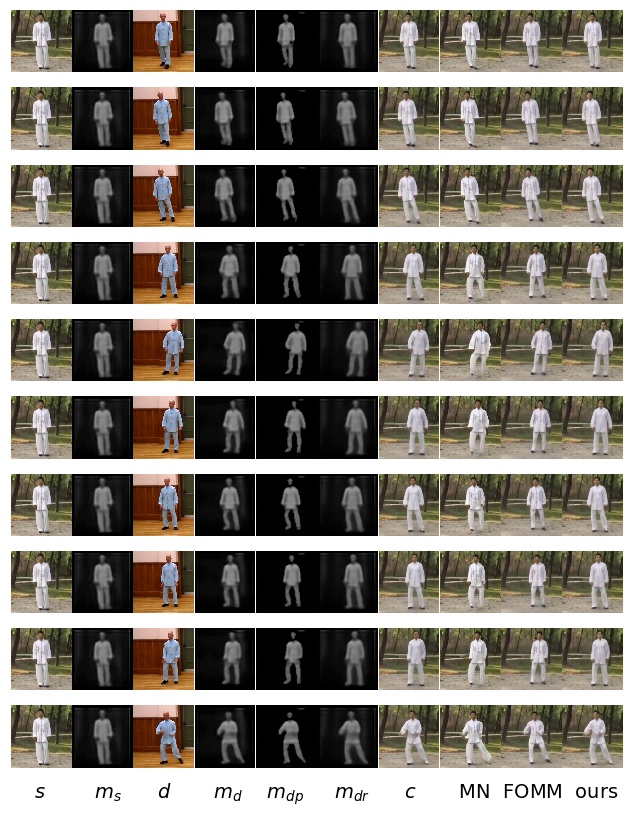}
\end{figure*}
\begin{figure*}[h!]
	\centering
    \ContinuedFloat
    \includegraphics[width=0.95\linewidth]{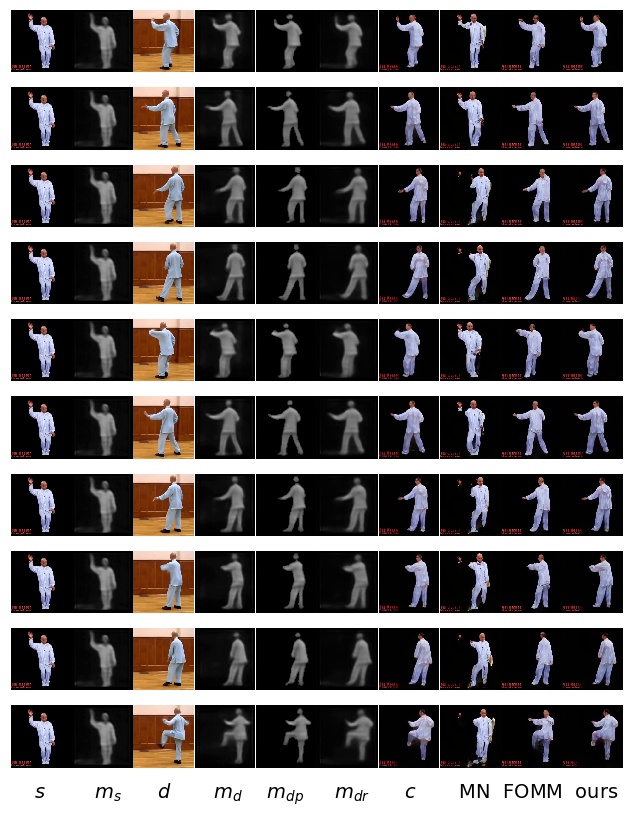}
\end{figure*}
\begin{figure*}[h!]
	\centering
    \ContinuedFloat
    \includegraphics[width=0.95\linewidth]{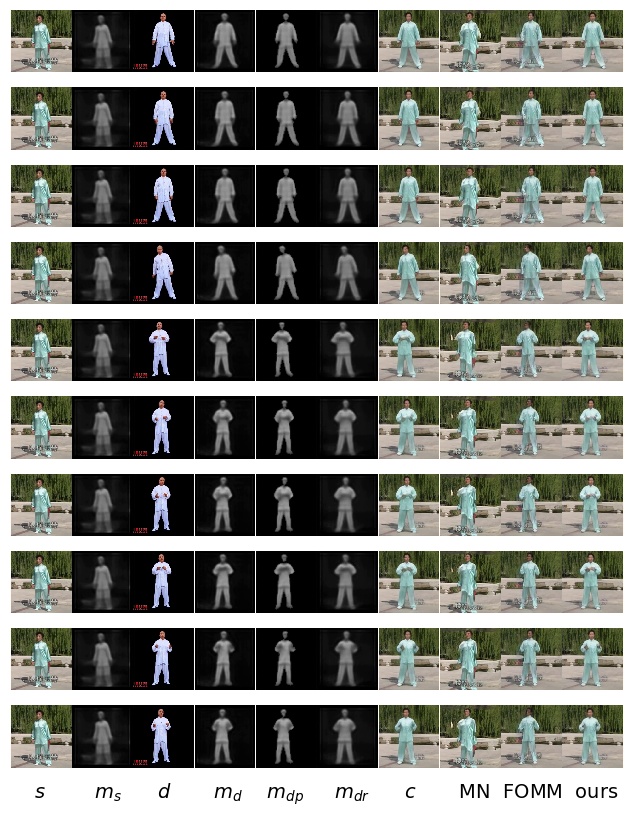}
\end{figure*}
\begin{figure*}[h!]
	\centering
    \ContinuedFloat
    \includegraphics[width=0.95\linewidth]{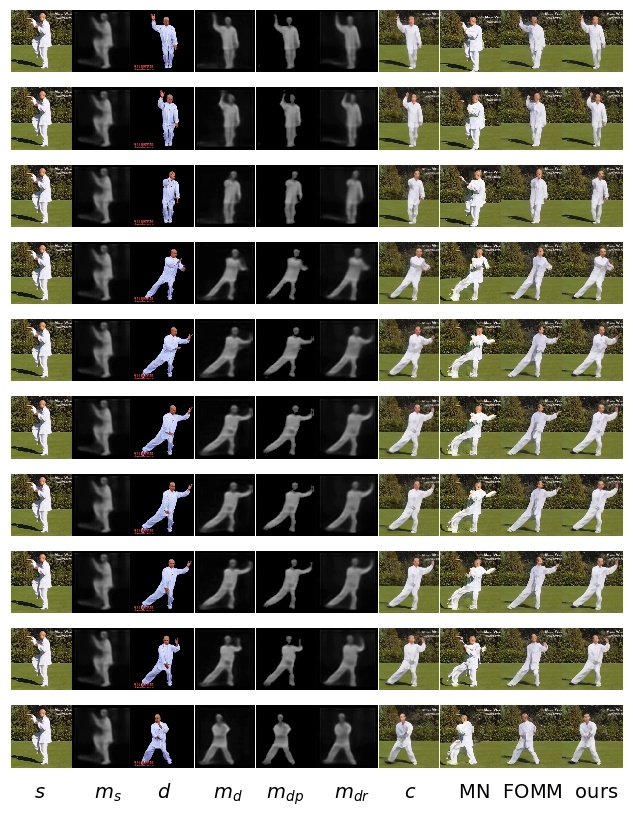}
\end{figure*}
\begin{figure*}[h!]
	\centering
    \ContinuedFloat
    \includegraphics[width=0.95\linewidth]{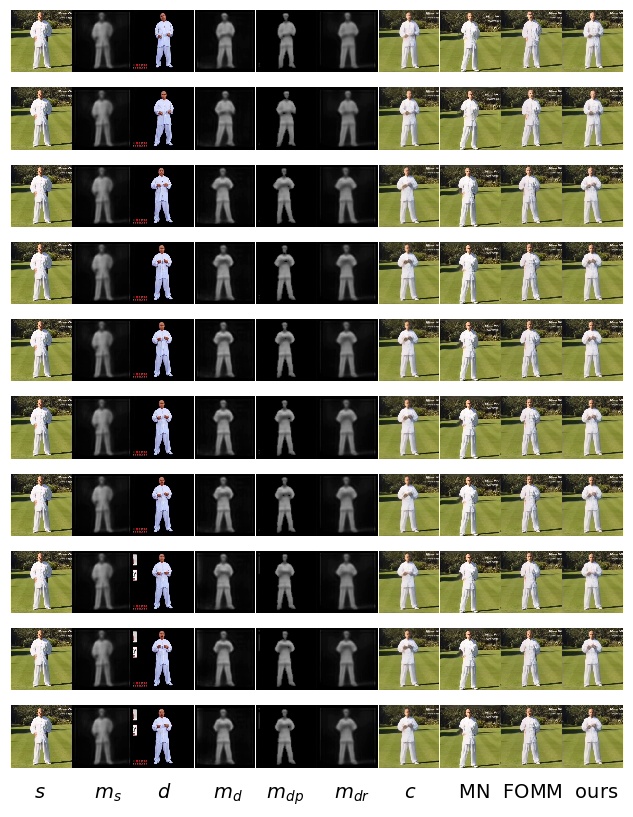}
\end{figure*}
\begin{figure*}[h!]
	\centering
    \ContinuedFloat
    \includegraphics[width=0.95\linewidth]{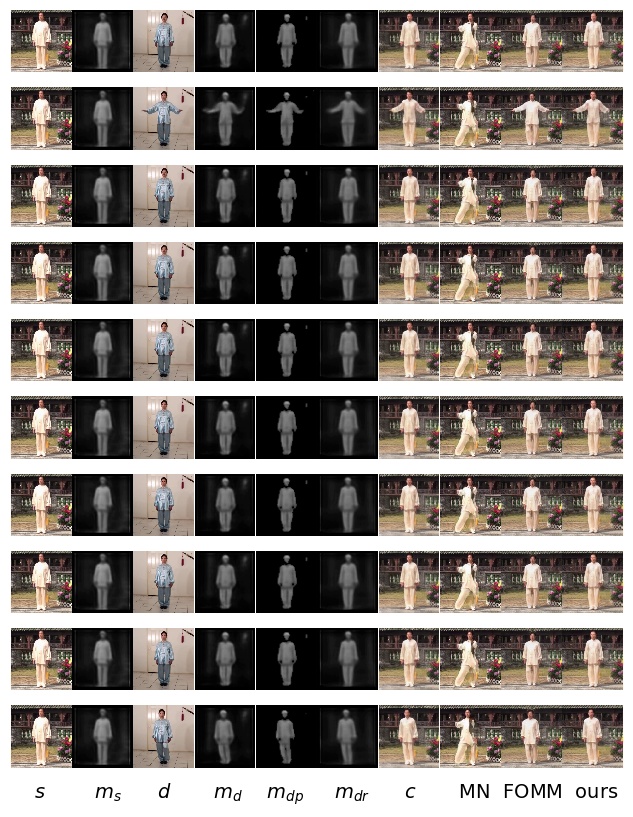}
\end{figure*}
\begin{figure*}[h!]
	\centering
    \ContinuedFloat
    \includegraphics[width=0.95\linewidth]{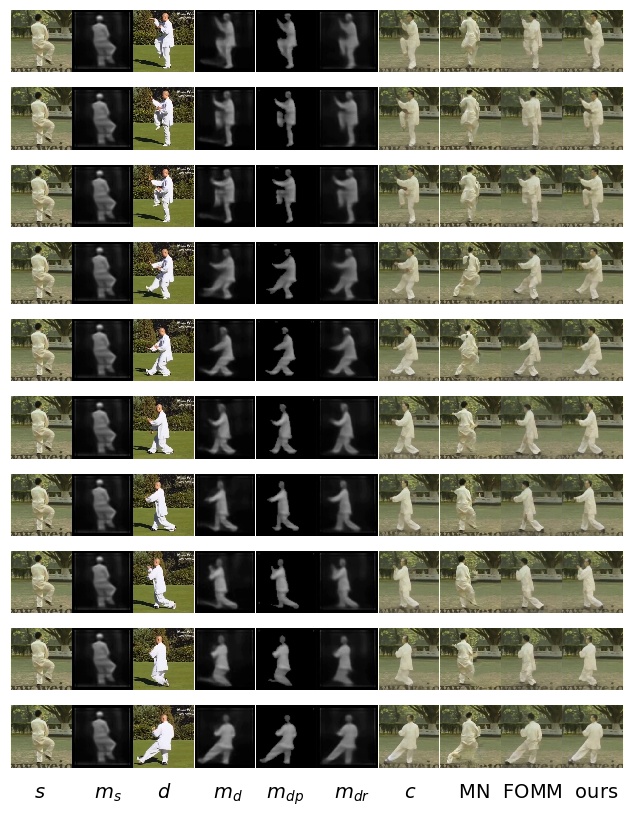}
\end{figure*}
\begin{figure*}[h!]
	\centering
    \ContinuedFloat
    \includegraphics[width=0.95\linewidth]{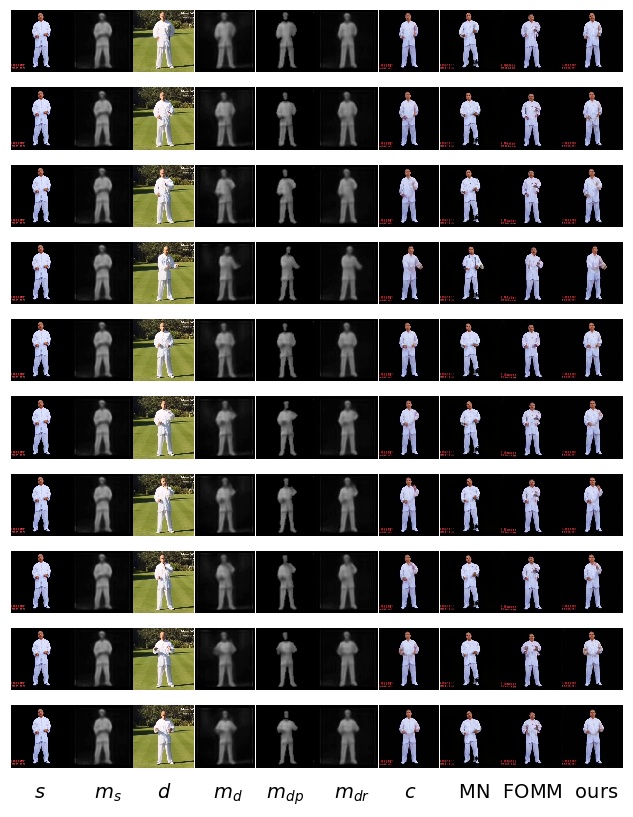}
\end{figure*}

\begin{figure*}[t]
	\centering
	\caption{Final and intermediate results generated by our method for BAIR, compared to the SOTA methods.}
    \includegraphics[width=0.95\linewidth]{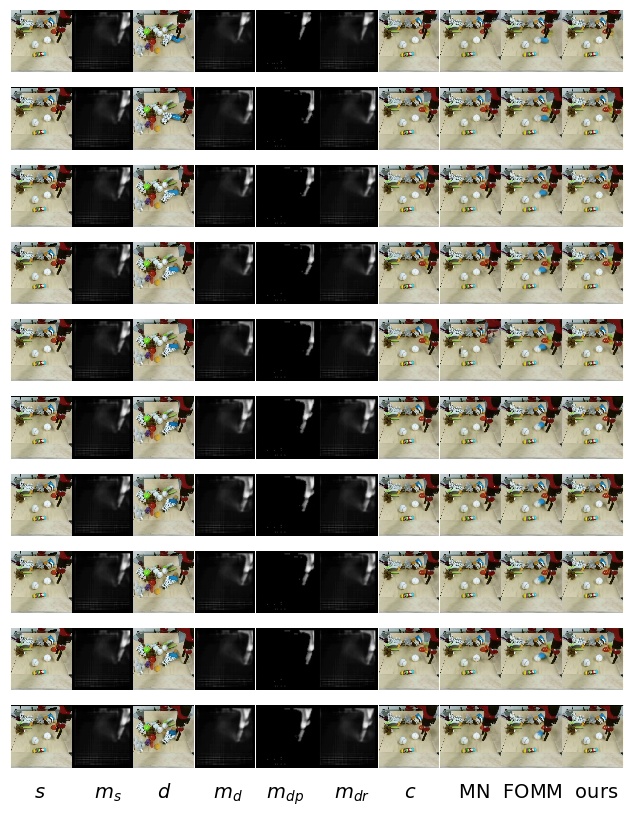}
        \label{fig:bair_examples}
\end{figure*}
\begin{figure*}[h!]
	\centering
    \ContinuedFloat
    \includegraphics[width=0.95\linewidth]{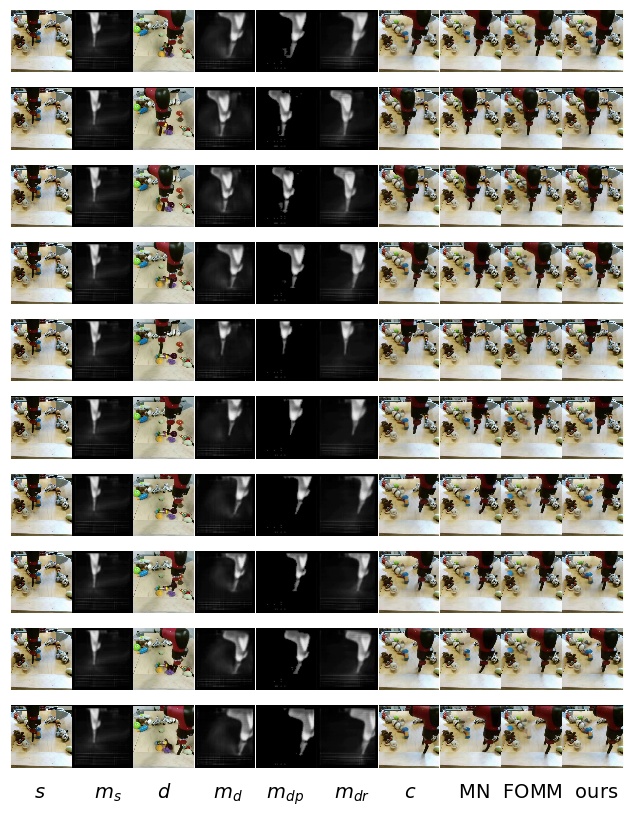}
\end{figure*}
\begin{figure*}[h!]
	\centering
    \ContinuedFloat
    \includegraphics[width=0.95\linewidth]{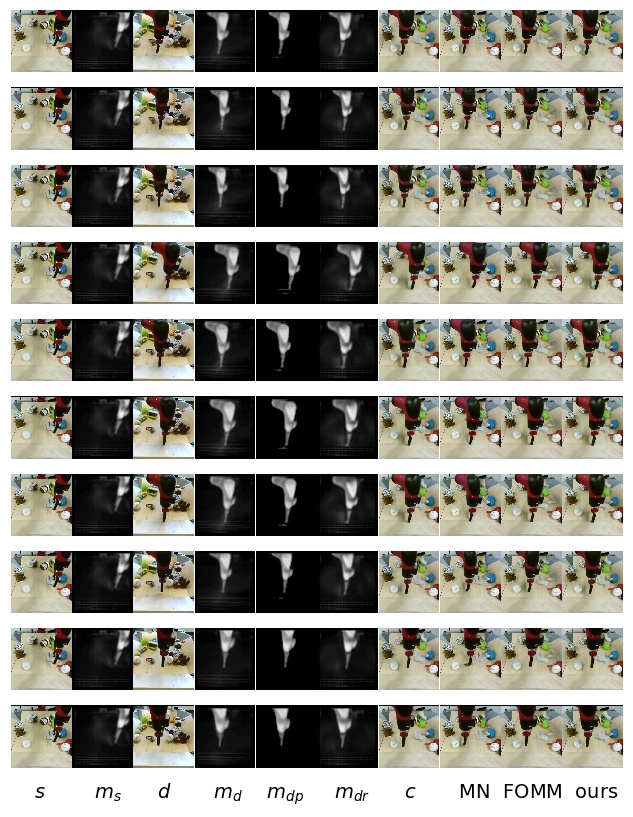}
\end{figure*}
\begin{figure*}[h!]
	\centering
    \ContinuedFloat
    \includegraphics[width=0.95\linewidth]{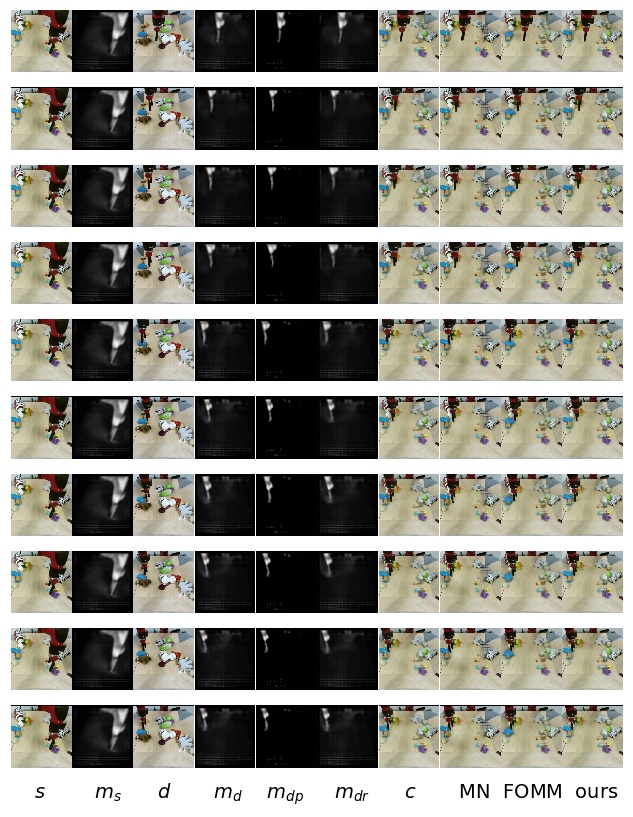}
\end{figure*}
\begin{figure*}[h!]
	\centering
    \ContinuedFloat
    \includegraphics[width=0.95\linewidth]{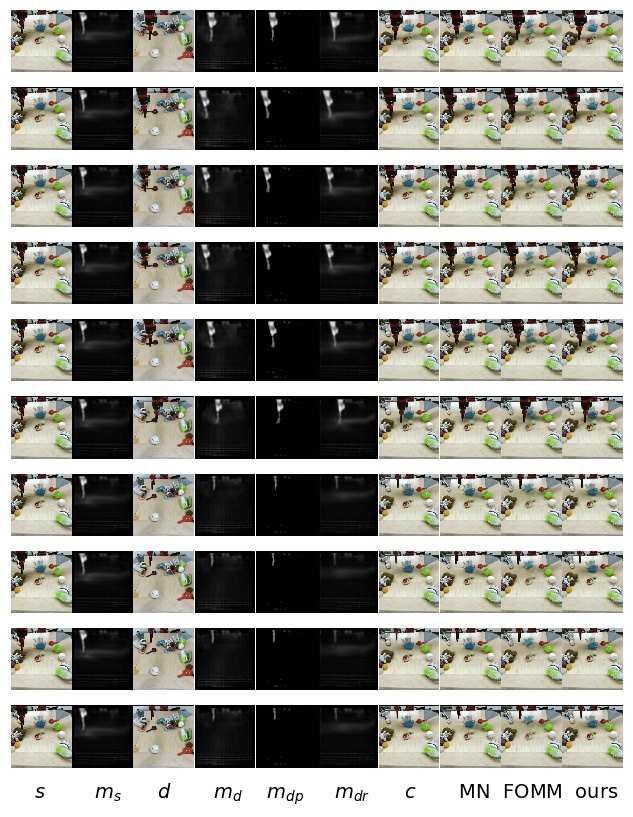}
\end{figure*}
\begin{figure*}[h!]
	\centering
    \ContinuedFloat
    \includegraphics[width=0.95\linewidth]{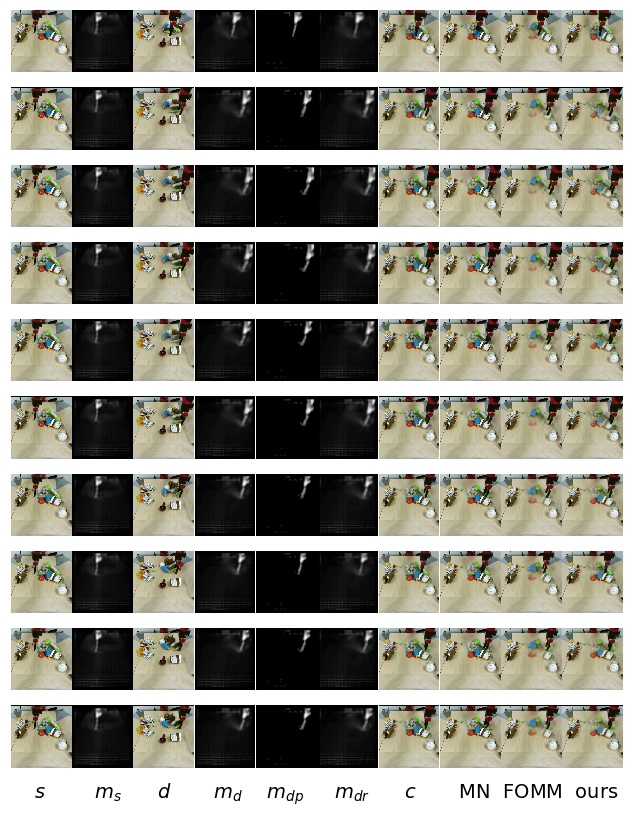}
\end{figure*}
\begin{figure*}[h!]
	\centering
    \ContinuedFloat
    \includegraphics[width=0.95\linewidth]{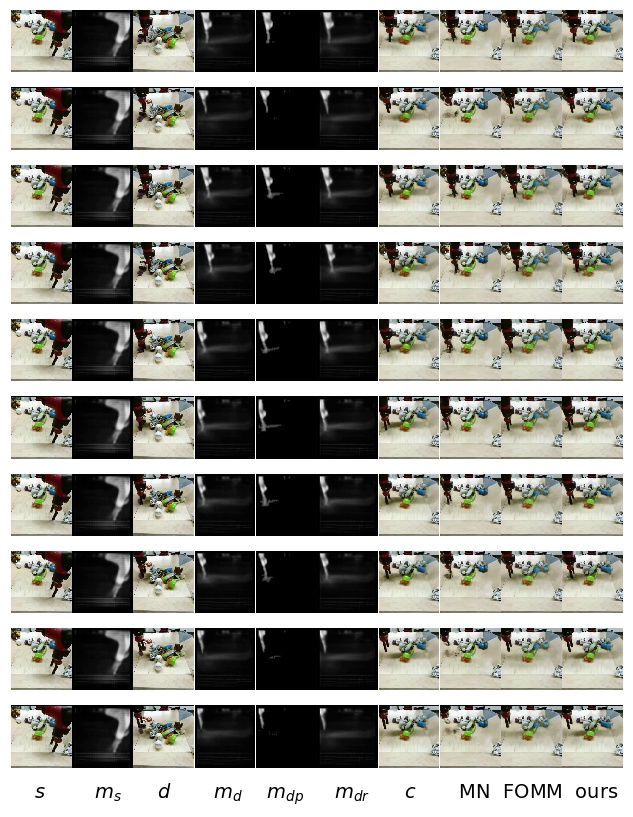}
\end{figure*}

\end{document}